\PassOptionsToPackage{table}{xcolor}
\documentclass[11pt]{article}
\usepackage[preprint]{acl}
\usepackage{times}
\usepackage{latexsym}
\usepackage[T1]{fontenc}
\usepackage[utf8]{inputenc}
\usepackage{microtype}
\usepackage{inconsolata}
\usepackage{graphicx}
\usepackage{booktabs}
\usepackage{multirow}
\usepackage{amsmath,amssymb}
\usepackage{xcolor}
\usepackage{subcaption}
\usepackage{enumitem}
\usepackage{longtable}
\usepackage{placeins}

\definecolor{ours}{HTML}{C0392B}
\definecolor{oursbg}{HTML}{FBEDEB}
\newcommand{\ours}{\textup{\textsc{LHA}}}

\newcommand{\accLHA}{76.6}

\newcommand{\accBase}{74.1}

\newcommand{\accEASE}{73.6}
\newcommand{\gainSFT}{1.8}

\newcommand{\gainBase}{2.5}

\newcommand{\gainEASE}{3.0}
\newcommand{\gainZwZ}{3.8}
\newcommand{\gainTreeVGR}{9.8}
\newcommand{\tokSFT}{20}

\newcommand{\gainQwen}{2.5}
\newcommand{\gainGreedy}{2.1}

\newcommand{\accAlSFT}{76.1}
\newcommand{\gainSFTonly}{1.4}
\newcommand{\accBzero}{75.7}
\newcommand{\tokBase}{24}

\newcommand{\sftSpatial}{4.0}

\newcommand{\sftFCP}{3.5}

\newcommand{\sftReas}{3.4}
\newcommand{\baseReas}{5.7}
\newcommand{\ciSFT}{[0.5, 2.9]}

\newcommand{\gainAttn}{0.9}

\newcommand{\rlStep}{50}
\newcommand{\ciGreedy}{[0.4, 4.0]}
\newcommand{\attnV}{1.8}
\newcommand{\attnTree}{1.6}

\newcommand{\routStart}{0.760}
\newcommand{\routStep}{50}
\newcommand{\routBzero}{0.761}
\newcommand{\routLHA}{0.860}
\newcommand{\routDiff}{0.10}
\newcommand{\routDiffCI}{[0.08, 0.12]}
\newcommand{\routAbove}{95}

\newcommand{\vstarStart}{.844}
\newcommand{\vstarBzero}{.842}
\newcommand{\vstarLHA}{.895}
\newcommand{\vstarMassStart}{0.076}
\newcommand{\vstarMassLHA}{0.126}
\newcommand{\vstarQueries}{1,343}

\renewcommand{\dbltopfraction}{0.9}

\title{Look at Where You Think: Localization-Head Alignment\\for Fine-Grained Perception in Thinking VLMs}

\author{%
  {\fontsize{12}{15}\selectfont\bfseries
    Yuan Wang\textsuperscript{1}\qquad
    Wei Wu\textsuperscript{2}\qquad
    Hongning Wang\textsuperscript{1,*}}\\[5pt]
  {\small\textsuperscript{1}Department of Computer Science and Technology, Tsinghua University}\\[2pt]
  {\small\textsuperscript{2}Ant Group}\\[3pt]
  {\fontsize{8.2}{10}\selectfont
    \href{mailto:wang-yua22@mails.tsinghua.edu.cn}{\texttt{wang-yua22@mails.tsinghua.edu.cn}}\quad
    \href{mailto:wuwei19850318@gmail.com}{\texttt{wuwei19850318@gmail.com}}\quad
    \href{mailto:hw-ai@tsinghua.edu.cn}{\texttt{hw-ai@tsinghua.edu.cn}}}\\[2pt]
  {\footnotesize *Corresponding author.}%
}

\begin{document}
\maketitle

\begin{abstract}
Thinking vision-language models reason for thousands of tokens, yet still miss small details in high-resolution images.
Zooming methods pay for extra inference passes, and outcome rewards never tell the model where to look at.
In Qwen3.5-4B, we find that the bottleneck is selection, not seeing: a few attention heads, which we call localization heads, pick the evidence out of thousands of visual tokens; ablating four of them costs 13.2 points more than ablating four random heads, and their focus fades as the reasoning grows.
We therefore propose Localization-Head Alignment (\ours{}), which makes attention a training signal: whenever the reasoning states a visual fact, these heads learn to look at the focal region rather than the background, as an auxiliary loss in both SFT and RL training.
On benchmarks of V*, HR-Bench 4K and TreeBench, \ours{} lifts Qwen3.5-4B from \accBase\% to \accLHA\% accuracy, the best single-pass 4B result we evaluate, ahead of multi-pass zooming and released 7B models.
Alignment alone adds \gainSFT{} points over otherwise identical training while shortening reasoning by \tokSFT\%, and the gains carry over to Qwen2.5-VL-3B.
\end{abstract}

\section{Introduction}

\begin{figure}[t]
  \centering
  \includegraphics[width=\columnwidth]{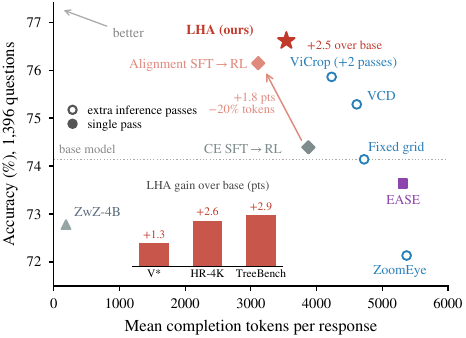}
  \caption{\textbf{Supervising where a few heads look gives the most accurate single-pass Qwen3.5-4B model, with shorter reasoning.} Overall accuracy on V*, HR-Bench 4K and TreeBench against mean generated tokens. Hollow markers use extra passes or crops. The arrow adds the alignment loss to otherwise identical SFT and RL; the inset breaks down LHA's gain over the base model by benchmark.}
  \label{fig:teaser}
\end{figure}

Thinking vision-language models (VLMs) reason for thousands of tokens before they answer~\citep{qwen35,vision-r1}, yet thinking longer does not make them see better: they still fail on fine-grained perception in high-resolution images~\citep{vstar,hrbench,treebench}.
For instance, ask such a model what color a kite is in a 4K photograph.
The image arrives as about 8,100 visual tokens, of which the kite covers only a handful, and by the time the reasoning gets to the kite, language priors~\citep{liu2025morethinking}, unrelated image regions and the model's own growing text all compete for its attention.
A model that looked at the right place early can drift away just when it needs to read the evidence.

Following the evidence from the pixels to the answer, we find that the bottleneck is selection, not seeing (\S\ref{sec:diagnosis}).
The vision encoder preserves small evidence; it is the thousands of surrounding tokens that bury it.
A few localization heads handle the selection: ablating just four of them in Qwen3.5-4B costs 13.2 points more than ablating four random heads.
They perceive the image at the tokens where the reasoning states a visual fact, not at the answer, and their preference for the evidence fades as the reasoning chain grows.
On V* for Qwen3.5-4B and in all 12 model--dataset settings we test across six direct-answer VLMs (Appendix~\ref{app:diag-routing}), correct answers are from the reasoning where  more attention is paid to the evidence than incorrect ones.

Yet no training signal speaks to these heads directly.
Cross-entropy imitates every output token alike, and outcome-reward RL such as GRPO~\citep{deepseekmath} credits a whole trajectory by its final answer; neither tells the heads where to look at while the model reasons.
Prior attention supervision targets other units: EASE fits a head-averaged map of one layer~\citep{ease}, Faithful-MR1 an inserted focus token~\citep{tian2026faithfulmr1} and AVAR the total visual attention~\citep{luo2026avar}; zooming methods add inference passes instead~\citep{vicrop,zoomeye,deepeyes}.

Our idea is to make attention itself the training target, placed exactly where the diagnosis says perception happens.
In particular, Localization-Head Alignment (\ours; Figure~\ref{fig:method}) supervises the localization heads, found from held-out grounding boxes and shown by ablation to matter, and it supervises them at the tokens where the reasoning states a fact about an annotated object.
For example, when the model writes ``the kite is blue'', we train the localization heads at the token that predicts ``kite'' to attend to the kite's box.
The loss contrasts the heads' attention on that evidence with an equally large background; the softmax normalizer cancels in the ratio, so it teaches where the heads look, not how much they look at the image.
One loss serves both stages, as an auxiliary objective on the model's own correct trajectories in SFT and on correct rollouts during GRPO, and inference is unchanged.

\ours{} is the most accurate single-pass Qwen3.5-4B model we evaluate, at \accLHA\% (Figure~\ref{fig:teaser}): it beats every inference-time intervention, including ViCrop with two extra passes per question, the attention-supervised RL method EASE by \gainEASE{} points, and released perception models up to 7B.
Both stages contribute.
On the same self-generated trajectories, alignment SFT adds \gainSFTonly{} points over cross-entropy SFT, and the gap grows to \gainSFT{} after identical outcome RL (95\% CI \ciSFT), most on questions that relate several objects.
In RL, the attention term adds \gainAttn{} points over the same RL without it, because, as a gradient decomposition shows, outcome rewards decide which trajectory to reinforce while the attention term consistently moves the heads toward the evidence.
\ours{} is also cheap: SFT updates 2.6\% of the language model, aligned policies write 19\% shorter rollouts so that each RL step runs 15\% faster, and the final model reasons with \tokBase\% fewer tokens than the base model.
The principle carries over to direct-answer Qwen2.5-VL-3B~\citep{qwen25vl} (\gainQwen{} points).

Our contributions are:
\begin{itemize}[leftmargin=*,itemsep=1pt,topsep=2pt]
  \item \textbf{Diagnosis.} In a thinking VLM the evidence survives encoding and is selected by a few causally important localization heads, which read the image where visual facts are stated and lose it as the chain grows (\S\ref{sec:diagnosis}).
  \item \textbf{Method.} \ours{} turns attention into an auxiliary objective for SFT and RL that, unlike prior attention supervision, acts on causally important heads at the fact tokens inside the reasoning, at no inference cost (\S\ref{sec:method}).
  \item \textbf{Evidence.} The most accurate single-pass Qwen3.5-4B model in our comparison, matched gains in accuracy on two backbones with shorter reasoning and faster RL training, and gradient and routing analyses showing that the attention term supplies the routing signal outcome rewards lack (\S\ref{sec:experiments}, \S\ref{sec:analysis}).
\end{itemize}

\section{Related Work}
\label{sec:related}

Thinking VLMs drift toward language priors as their chains grow~\citep{liu2025morethinking,visualswap}.
Inference-time methods recover small details with guided search, attention-selected crops or learned crop tools, at the cost of extra passes~\citep{vstar,zoomeye,vicrop,deepeyes}.
Perception RL rewards correct answers, grounded boxes or stated perception~\citep{vision-r1,treebench,wang2026moca}, and these signals act on the policy's outputs.
The closest methods supervise attention during RL.
EASE matches a head-averaged map of one layer to a smoothed evidence prior~\citep{ease}.
Faithful-MR1 supervises an inserted focus token on a prescribed set of layer--head pairs before a counterfactual attention reward~\citep{tian2026faithfulmr1}.
AVAR rewards visual attention after an attention-guided cold start~\citep{luo2026avar}.

Functional heads that retrieve context~\citep{retrieval-heads} or ground image regions~\citep{kang2025heads} have mainly been used for analysis and for training-free grounding or cropping~\citep{vicrop}.

Relative to EASE's layer-averaged maps, Faithful-MR1's inserted focus token and AVAR's visual-attention reward, \ours{} supervises independently selected localization heads at the native queries that predict visual facts.
Its contrast between the evidence and an equally large background cancels the attention softmax normalizer and is neutral under uniform attention, giving a direct regional-preference objective that alignment SFT and attention RL share without changing inference.
Appendix~\ref{app:related} discusses each line of work in detail.

\section{Setup}
\label{sec:setup}

\paragraph{Model.}
We study Qwen3.5-4B~\citep{qwen35} in thinking mode.
Its language model interleaves Gated DeltaNet linear-attention layers with eight full-attention layers (3, 7, \dots, 31) of 16 query heads each.
Because linear-attention layers compress the context into a fixed-size state, we study the 128 full-attention heads, which attend to every visual token directly.

\paragraph{Benchmarks and protocol.}
We evaluate on 1,396 multiple-choice questions about small evidence in high-resolution images: V*~\citep[191;][]{vstar}, HR-Bench 4K~\citep[800, 200 questions in four option orders;][]{hrbench} and TreeBench~\citep[405;][]{treebench}.
Qwen3.5-4B models are decoded in thinking mode at temperature 1 with a 32K-token budget.
Released models from other backbones use their own generation format.
Rules extract the answers, and an LLM judge (Qwen3.6-35B-A3B) reads only the responses whose committed option the rules cannot parse.
A response is \emph{budget-truncated} if it reaches the 32K budget without finishing.
The protocol and the base-model reference were frozen before the experiments, and every model in Table~\ref{tab:main} is scored by the same rules and judge.
We checked the scoring of all reported evaluations by hand, including every judge decision, and it agreed with the manual labels throughout (Appendix~\ref{app:protocol}).
RL-trained Qwen3.5-4B models are reported as the mean of two independent decodes where both are available (single-decode rows are marked); paired bootstrap intervals for all controlled comparisons are collected in Appendix~\ref{app:protocol}.

\section{Where Thinking VLMs Select the Evidence}
\label{sec:diagnosis}

\begin{figure*}[t]
  \centering
  \includegraphics[width=\textwidth]{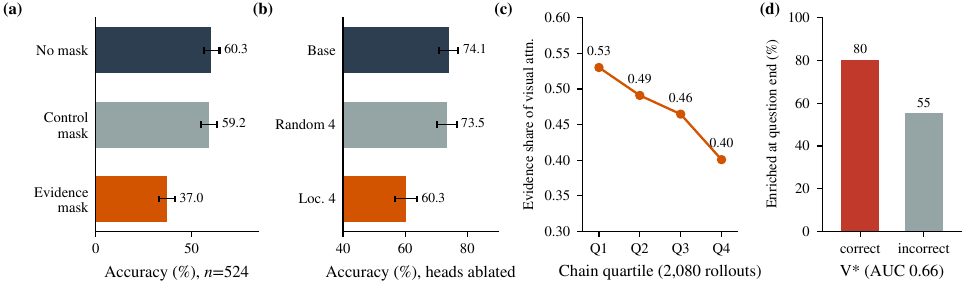}
  \caption{\textbf{Thinking VLMs need small evidence and rely on a few localization heads to select it} (Qwen3.5-4B). (a) Masking the evidence hurts far more than masking an equal-area region elsewhere. (b) Ablating four localization heads hurts far more than ablating four random heads of the same layers. (c) The heads' share of visual attention inside the evidence box declines along the chain of thought. (d) On V*, the heads concentrate on the evidence at the end of the question far more often when the answer is correct. Error bars: 95\% bootstrap CIs.}
  \label{fig:diagnosis}
\end{figure*}

Where does perception break in a thinking VLM?
We follow the evidence from the pixels to the answer.
Evidence masking, post-encoder token controls and head interventions point to a concrete supervision target, namely a few localization heads at the positions that predict visual facts.

\subsection{The answer hinges on small evidence}
\label{sec:diag-evidence}
Masking the ground-truth region of a question costs Qwen3.5-4B 22.1 points more than masking an equal-area region elsewhere (Figure~\ref{fig:diagnosis}a), and the smaller the evidence, the larger the cost (Appendix~\ref{app:diag-mask}).
Fine-grained perception is therefore a search for a few relevant tokens among thousands.

\subsection{The evidence survives encoding but is buried}
\label{sec:diag-tokens}
Is small evidence lost in the vision encoder or among the visual tokens that follow it?
We edit the visual tokens after encoding, so the encoder still sees the whole image.
Given only the tokens inside the evidence box, a small subset of what it already receives, Qwen3.5-4B answers V* and TreeBench questions as accurately as from the full image (60.7\% vs.\ 60.1\%).
Given an equal-area set of other tokens instead, it falls to 38.0\%.
Appending the box tokens to the full image adds almost nothing, and on synthetic small-object scenes the box tokens alone even raise Qwen2.5-VL-3B accuracy by 16.8 points (Appendix~\ref{app:diag-tokens}).
The encoder preserves the evidence, and more of it does not help; what the model must do is pick it out from the tokens around it.

\subsection{A few localization heads select the evidence}
\label{sec:diag-heads}
Following retrieval-head scoring~\citep{retrieval-heads}, we rank the 128 full-attention heads of Qwen3.5-4B by how strongly they attend to the evidence box.
The ranking uses held-out grounding questions, never correctness; computed independently on V*, it has Spearman correlation 0.80 with this ranking and shares six of the top eight heads (Appendix~\ref{app:heads}).
Zeroing only the top four heads costs 13.2 points more than zeroing four random heads of the same layers, with large drops on every benchmark (Figure~\ref{fig:diagnosis}b).
Such heads are known from short-answer grounding~\citep{kang2025heads,gandikota2026gazeheads,vicrop}.
In a thinking model, they matter for the answer itself, not only for the reasoning that precedes it (Appendix~\ref{app:diag-ablation}): in a forced-choice probe without any reasoning, ablating eight probe-selected heads lowers accuracy from 52.6\% to 36.8\%, while layer-matched random heads leave it at 52.6\%.

\subsection{Evidence routing predicts correctness and fades during reasoning}
\label{sec:diag-answer}
On V*, where a single small object decides the answer, the routing of these heads at the end of the question already separates correct from incorrect answers (Figure~\ref{fig:diagnosis}d).
TreeBench questions relate several objects that the reasoning visits one by one, so their evidence is read inside the chain.

Inside the reasoning, the heads read the image at the tokens that state visual facts, where they clearly prefer the stated object over an equal-area background (Appendix~\ref{app:diag-routing}).
This reading weakens as the chain grows: the heads' share of visual attention on the evidence falls by a quarter from the beginning of the chain to its end (Figure~\ref{fig:diagnosis}c).
The same decline appears on V* and TreeBench rollouts, where the heads' preference for the evidence over an equal-size background drops steadily from the first tenth of the reasoning to the last (Appendix~\ref{app:diag-routing}).
Restoring only the amount of attention that reaches the image does not help (Appendix~\ref{app:diag-decay}).
The heads need to look at the right place, not more at the image.

\paragraph{Takeaway.}
The evidence is in the input, and the model has heads that can find it.
They attend to it at fact tokens rather than at the answer, and less sharply as the reasoning chain grows.
An outcome reward sees only the final answer, so \ours{} places regional supervision where visual facts are generated.

\section{Localization-Head Alignment}
\label{sec:method}

\begin{figure*}[t]
  \centering
  \includegraphics[width=\textwidth]{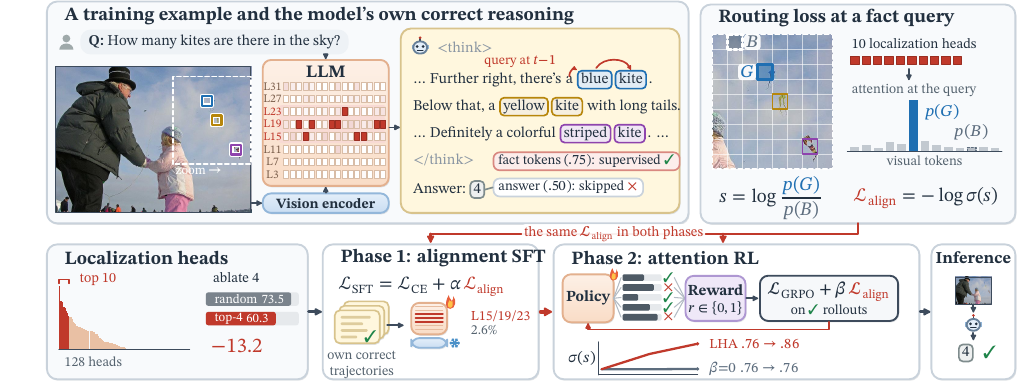}
  \caption{\textbf{\ours{} on a TallyQA training question.} \emph{Top:} in the model's own correct reasoning, each fact phrase names an annotated object (same color in the image). At the query $t{-}1$ that predicts a fact token, the routing loss raises the attention of the localization heads (red cells in the LLM) on the object's visual tokens $G$ against an equally large background $B$; the answer token, where the preference is already at chance, is skipped (.75 vs .50 on average). \emph{Bottom:} the heads are the top ten of 128, ranked on grounding questions; ablating the top four costs 13.2 points more than ablating four random heads. The same loss joins cross-entropy in alignment SFT (flame: trained; snowflake: frozen) and GRPO on correct rollouts in attention RL, where it lifts the heads' preference $\sigma(s)$ from .76 to .86 in 50 steps (Table~\ref{tab:rl-mechanism}b). Inference is unchanged.}
  \label{fig:method}
\end{figure*}

The diagnosis tells us to supervise a few localization heads at the positions where visual facts are read.
\ours{} turns this into one routing loss used in two phases (Figure~\ref{fig:method}): alignment SFT on the model's own correct trajectories, then attention RL, i.e., outcome-reward RL that keeps the same loss on correct rollouts.
We describe it for the thinking Qwen3.5-4B; an adapted version for the direct-answer Qwen2.5-VL-3B supervises the answer position (Appendix~\ref{app:impl}).

\subsection{Routing loss on localization heads}
\label{sec:loss}
For an image $x$ and question $q$, let $y=(y_1,\dots,y_T)$ be a trajectory of the model and $\mathcal{V}$ the positions of the visual tokens in the input.
For head $h$ with attention $a_{h,i}$ at query position $i$ and a set of visual tokens $R\subseteq\mathcal{V}$, let
\begin{equation}
  p_{h,i}(R)=\frac{\sum_{v\in R}a_{h,i}(v)}{\sum_{v\in\mathcal{V}}a_{h,i}(v)}
  \label{eq:share}
\end{equation}
be the share of the head's visual attention that falls in $R$.
Normalizing within the image discards attention on text and sink tokens~\citep{xiao2024streamingllm,kang2025seewhat}, which grows with the chain (\S\ref{sec:diag-answer}).

\paragraph{Heads.}
$\mathcal{H}$ contains the top-scoring localization heads of \S\ref{sec:diag-heads}.
For Qwen3.5-4B these are the ten top heads, which lie in layers 15, 19 and 23 and include the four ablated heads.

\paragraph{Positions.}
A thinking model reads the image at fact tokens, such as a stated count or a word of an annotated object phrase.
There the supervised heads prefer the evidence over an equal-size background, $p(G)/(p(G){+}p(B))\approx0.75$, whereas at the final answer token the preference is at chance (0.50), because the answer follows from facts already stated (Appendix~\ref{app:diag-routing}).
We therefore supervise the facts, not the answer; a direct-answer model reads the image at the answer and is supervised there.
For each supervised token $y_t$ we take the query at position $t{-}1$, whose output predicts $y_t$, and collect these positions in $\mathcal{F}(y)$.
Each $i\in\mathcal{F}(y)$ receives an evidence set $G_i\subseteq\mathcal{V}$, the visual tokens covered by the annotated boxes the token refers to, and a background $B_i$ of the same size drawn from the visual tokens outside every box.

\paragraph{Loss.}
We score how strongly head $h$ prefers the evidence at position $i$ by the log-odds
\begin{equation}
  s_{h,i}=\log\frac{p_{h,i}(G_i)}{p_{h,i}(B_i)},
  \label{eq:score}
\end{equation}
and penalize it with a logistic loss, averaged over a set of positions $\mathcal{P}$:
\begin{equation}
  \mathcal{L}_{\text{align}}(y;\mathcal{P})=-\frac{1}{|\mathcal{P}||\mathcal{H}|}\sum_{i\in\mathcal{P},\,h\in\mathcal{H}}\log\sigma(s_{h,i}).
  \label{eq:align}
\end{equation}
Since $\sigma(s_{h,i})=p_{h,i}(G_i)/(p_{h,i}(G_i)+p_{h,i}(B_i))$, minimizing the loss moves attention from the background to the evidence.
Two properties make the loss target where the head looks.
The softmax normalizer, and with it the denominator of Eq.~\eqref{eq:share}, cancels in $s_{h,i}$, so the loss depends only on the attention logits of evidence and background tokens; attention to text, sinks and the rest of the image does not enter it.
Moreover, because $B_i$ is as large as $G_i$, uniform attention scores a neutral $\tfrac12$, so the loss measures per-token preference, not box size.

\paragraph{From diagnosis to design.}
Each choice comes from a diagnostic finding.
The heads are found from grounding boxes alone, without looking at correctness, and the model causally relies on them (\S\ref{sec:diag-heads}).
The positions are where these heads read the image; the answer token, where their preference is at chance, is left alone.
Box-matched fact tokens also concentrate the signal: only 13.7\% of 64 positions sampled uniformly from a trajectory fall on fact content (Appendix~\ref{app:training}).
The contrast stays inside the image because answers use context around the box and because restoring visual mass alone does not help (\S\ref{sec:diag-answer}): the loss should move attention within the image, not add more.

\subsection{Phase 1: alignment SFT}
\label{sec:align}
The first phase trains on correct trajectories $y$ with
\begin{equation}
  \mathcal{L}_{\text{SFT}}=\mathcal{L}_{\text{CE}}(y)+\alpha\,\mathcal{L}_{\text{align}}\bigl(y;\mathcal{F}(y)\bigr),
  \label{eq:sft}
\end{equation}
where $\mathcal{L}_{\text{CE}}$ is the token-averaged cross-entropy of the whole trajectory.
The trajectories are the base model's own correct best-of-five samples on box-annotated counting and spatial questions from TallyQA~\citep{tallyqa}, PixMo-Count~\citep{molmo}, Spatial-MQA~\citep{spatialmqa} and VSR~\citep{vsr}, so no external teacher is involved.
We train only the attention projections of the supervised layers (2.6\% of the language-model parameters), and the alignment gradient reaches only the query projections of the supervised heads (Appendix~\ref{app:training}).
Because the raw gradients differ by orders of magnitude, $\alpha$ is set at every update from the batch gradients:
\begin{equation}
  \alpha=\rho\,f\,\frac{\lVert g_{\text{CE}}^{\,QK}\rVert_2}{\lVert g_{\text{align}}\rVert_2},
  \label{eq:alpha}
\end{equation}
where $g_{\text{CE}}^{\,QK}$ is the cross-entropy gradient on the query and key projections, $f$ the fraction of trajectories in the batch with $\mathcal{F}(y)\neq\emptyset$, and $\rho$ the dose.
The update uses $g_{\text{CE}}+\alpha g_{\text{align}}$, clipped to norm 1.

\begin{table*}[t]
\centering
\small
\setlength{\tabcolsep}{3.6pt}
\begin{tabular}{@{}lccccccccccr@{}}
\toprule
\multirow{2}{*}{Method} & \multicolumn{3}{c}{V$^*$} & \multicolumn{3}{c}{HR-Bench 4K} & \multicolumn{3}{c}{TreeBench} & \multirow{2}{*}{\textbf{Overall}} & \multirow{2}{*}{Tokens} \\
\cmidrule(lr){2-4} \cmidrule(lr){5-7} \cmidrule(lr){8-10}
 & Attr. & Spatial & All & FSP & FCP & All & Perc. & Reas. & All & & \\
\midrule
\multicolumn{12}{@{}l}{\textit{Released perception models (own backbones)}} \\[1pt]
\quad ZwZ-4B\textsuperscript{\dag} & 90.4 & 88.2 & 89.5 & 90.0 & 72.2 & 81.1 & 63.8 & 39.5 & 48.4 & 72.8 & 185 \\
\quad DeepEyes-7B (no tools)\textsuperscript{\dag} & 73.0 & 68.4 & 71.2 & 85.8 & 61.0 & 73.4 & 57.0 & 28.9 & 39.3 & 63.2 & 45 \\
\quad TreeVGR-7B\textsuperscript{\dag} & 82.6 & 84.2 & 83.2 & 87.8 & 58.8 & 73.2 & 61.1 & 37.5 & 46.2 & 66.8 & 60 \\
\midrule
Qwen3.5-4B (base)\textsuperscript{\dag} & 84.3 & 85.5 & 84.8 & 91.0 & 77.0 & 84.0 & 73.8 & 35.5 & 49.6 & 74.1 & 4,678 \\
\midrule
\multicolumn{12}{@{}l}{\textit{Inference-time interventions on Qwen3.5-4B (extra passes)}} \\[1pt]
\quad ViCrop & 83.5 & 86.8 & 84.8 & 91.9 & 79.0 & 85.4 & 72.1 & 41.4 & 52.7 & 75.9 & 4,230 \\
\quad VCD\textsuperscript{\dag} & 82.6 & 93.4 & 86.9 & 90.5 & 80.0 & 85.2 & 71.1 & 37.9 & 50.1 & 75.3 & 4,612 \\
\quad ZoomEye\textsuperscript{\dag} & 87.0 & 97.4 & 91.1 & 89.5 & 69.0 & 79.2 & 69.8 & 37.1 & 49.1 & 72.1 & 5,371 \\
\quad Fixed-grid zoom\textsuperscript{\dag} & 82.6 & 90.8 & 85.9 & 90.8 & 76.2 & 83.5 & 69.1 & 39.1 & 50.1 & 74.1 & 4,725 \\
\midrule
\multicolumn{12}{@{}l}{\textit{Training on Qwen3.5-4B (single pass)}} \\[1pt]
\quad EASE & 82.6 & 86.2 & 84.0 & 90.1 & 77.2 & 83.7 & 70.8 & 36.1 & 48.9 & 73.6 & 5,316 \\ \addlinespace[2pt]
\quad CE SFT\textsuperscript{\dag} & 82.6 & 80.3 & 81.7 & 91.0 & 78.5 & 84.8 & 73.2 & 39.1 & 51.6 & 74.7 & 3,730 \\
\quad Alignment SFT\textsuperscript{\dag} & 86.1 & 85.5 & 85.9 & \textbf{92.8} & 78.0 & 85.4 & 73.2 & \textbf{41.4} & \textbf{53.1} & 76.1 & 3,940 \\ \addlinespace[2pt]
\quad CE SFT $\rightarrow$ RL & 84.3 & 84.2 & 84.3 & 90.9 & 77.5 & 84.2 & \textbf{75.2} & 35.9 & 50.4 & 74.4 & 3,876 \\
\quad Alignment SFT $\rightarrow$ RL & \textbf{87.0} & \textbf{88.2} & \textbf{87.4} & 90.8 & \textbf{81.0} & 85.9 & 72.8 & 39.3 & 51.6 & 76.1 & \textbf{3,112} \\ \addlinespace[2pt]
\rowcolor{oursbg}\quad \textbf{\ours{}} (attention RL) & 85.7 & 86.8 & 86.1 & 92.1 & \textbf{81.0} & \textbf{86.6} & 71.8 & 41.2 & 52.5 & \textbf{76.6} & 3,541 \\
\bottomrule
\end{tabular}
\caption{\textbf{Supervising where a few heads look gives the most accurate single-pass Qwen3.5-4B model.} Accuracy (\%) in thinking mode (temperature 1, 32K-token budget) on V$^*$ (attribute and spatial questions), HR-Bench 4K (single- and cross-instance) and TreeBench (perception and reasoning); Overall pools all 1,396 questions and Tokens is the mean completion length. Below EASE, SFT on the same self-generated trajectories uses cross-entropy (CE) alone or adds the alignment loss, and each is followed by the same outcome RL; \ours{} continues alignment SFT with attention RL. Rows average two decodes (\dag: one). \textbf{Bold}: best single-pass Qwen3.5-4B model.}
\label{tab:main}
\end{table*}

\subsection{Phase 2: attention-supervised RL}
\label{sec:rl}
The second phase starts from the aligned model and adds the same loss to outcome-reward RL.
For each question it samples a group of $K$ rollouts $\{y^k\}$ with rewards $r^k\in\{0,1\}$ (1 for the correct committed option) and minimizes
\begin{equation}
  \mathcal{L}_{\text{RL}}=\mathcal{L}_{\text{GRPO}}+\frac{\beta}{|\mathcal{Y}|}\sum_{y\in\mathcal{Y}^{+}}\mathcal{L}_{\text{align}}\bigl(y;\mathcal{F}(y)\bigr),
  \label{eq:rl}
\end{equation}
where $\mathcal{L}_{\text{GRPO}}$ is the clipped group-relative policy loss with a KL penalty to the starting model~\citep{deepseekmath} and $\mathcal{Y}^{+}$ the correct rollouts among the rollouts $\mathcal{Y}$ of an optimizer mini-batch.
The reward decides which trajectories to reinforce; the attention term decides where the localization heads look inside them.
Only correct rollouts are supervised, because their reasoning reached the annotated answer.
Their fact tokens are the object nouns of the annotated phrases and the count tokens that enumerate instances (``person 2'') in the thinking span.
Unlike in SFT, where each fact token has its own $G_i$, all fact queries of a rollout share one evidence set $G$, the union of the annotated boxes the rollout refers to, and the gradient flows through all parameters (Appendix~\ref{app:training}).

\paragraph{Configuration.}
\ours{} uses $\rho{=}0.1$ in alignment SFT and $\beta{=}0.01$ in attention RL, which trains on questions that the base model answers correctly in some but not all samples (Appendix~\ref{app:rl}).

\section{Experiments}
\label{sec:experiments}

\subsection{Baselines and controls}
\textbf{Released perception models}, evaluated under our protocol on their own backbones: ZwZ-4B~\citep{zwz} on Qwen3-VL-4B, and DeepEyes-7B~\citep{deepeyes} and TreeVGR-7B~\citep{treebench} on Qwen2.5-VL-7B~\citep{qwen25vl} (NoisyRollout-7B~\citep{noisyrollout} in Appendix~\ref{app:endpoints}).
\textbf{Inference-time interventions} on Qwen3.5-4B, which buy accuracy with extra inference passes (Appendix~\ref{app:cost}): ViCrop~\citep{vicrop} (relative-attention crop added as a second image), VCD~\citep{vcd} (contrastive decoding with a distorted image), ZoomEye~\citep{zoomeye} (confidence-guided tree search over sub-images) and a fixed-grid crop.
\textbf{Attention-supervised RL}: EASE~\citep{ease}, which adds a KL term toward a Gaussian evidence prior on a head-averaged attention map during DAPO~\citep{dapo}.
Outcome RL from the base model with GRPO~\citep{deepseekmath} or DAPO stays within decoding noise of the base model (Appendix~\ref{app:endpoints}).

\textbf{Our training rows.}
\textbf{CE SFT} and \textbf{alignment SFT} share trajectories, data order, updates and trainable parameters and differ only in the alignment loss ($\alpha$); each is then followed by the same outcome RL.
\ours{} continues alignment SFT with attention RL, and its $\beta{=}0$ control runs the same RL from the same checkpoint without the attention term (Appendix~\ref{app:rl}; paired intervals in Table~\ref{tab:stats}).

\subsection{Attention supervision makes perception more accurate}
\label{sec:main-results}
\ours{} is the most accurate single-pass Qwen3.5-4B model in Table~\ref{tab:main}, at \accLHA\% (\gainBase{} points above the base model with \tokBase\% fewer tokens), and it outperforms every inference-time intervention, including ViCrop, which runs two extra attention passes and a crop on every question.

\paragraph{Alignment makes SFT more accurate, and the gain survives RL.}
Alignment SFT alone reaches \accAlSFT\%, \gainSFTonly{} points above CE SFT on the same trajectories and on par with the best inference-time intervention at a single pass.
After identical outcome RL the gap grows to \gainSFT{} points (95\% CI \ciSFT), and the aligned model uses \tokSFT\% fewer completion tokens.
The gain holds under greedy decoding (\gainGreedy{} points, 95\% CI \ciGreedy; Appendix~\ref{app:t0}), and it is not a by-product of fewer truncated responses: on the 2,713 question--decode pairs where both models finish, the aligned pipeline answers 43 more questions correctly, and these pairs account for 86.8\% of its token saving (Appendix~\ref{app:efficiency}).
It is largest on questions that combine several pieces of evidence: \sftSpatial{} points on the spatial questions of V*, \sftFCP{} on the cross-instance questions of HR-Bench and \sftReas{} on the reasoning questions of TreeBench.
Aligning just 44 fact tokens thus leaves a mark that outlasts a full RL stage without the loss.

\paragraph{Attention RL adds a direct routing signal.}
With the same RL, the outcome reward alone leaves the aligned model where SFT put it (\accBzero\% for the $\beta{=}0$ control), whereas keeping the loss lifts it to \accLHA\%: \gainAttn{} points above the control in each of the two decodes, with \attnV{} on V* and \attnTree{} on TreeBench.
Meanwhile, the attention term sharpens the heads' evidence preference, while the control's stays flat
(\S\ref{sec:analysis}).
\ours{}'s largest gain over the base model is \baseReas{} points on TreeBench reasoning.

\paragraph{Comparison with attention-supervised RL.}
EASE, the closest attention-supervised RL method, reaches \accEASE\% on our backbone, \gainEASE{} points below \ours{}.
Both supervise attention toward the evidence, but at different places: EASE matches a layer-averaged map to a smoothed prior, whereas \ours{} supervises the localization heads at the fact queries, with a normalizer-free contrast.

\paragraph{Comparison with released perception models.}
Under the same protocol, the 4B \ours{} model is more accurate overall than every released perception model we evaluate, including those built on 7B backbones.
It leads ZwZ-4B by \gainZwZ{} points and TreeVGR-7B by \gainTreeVGR{}, and it leads both on HR-Bench 4K and TreeBench, where evidence is tiny or spread over several objects.

\begin{table}[t]
\centering
\small
\setlength{\tabcolsep}{4pt}
\begin{tabular}{@{}lrrr@{}}
\toprule
Method & V$^*$ & Tree & V$^*$+Tree \\
\midrule
Base & 74.3 & 39.8 & 50.8 \\
CE SFT & 77.0 & 39.8 & 51.7 \\
Alignment SFT & 78.0 & 40.5 & 52.5 \\
\midrule
CE SFT $\rightarrow$ RL & 75.4 & 38.8 & 50.5 \\
Alignment SFT $\rightarrow$ RL & \textbf{78.5} & \textbf{41.0} & \textbf{53.0} \\
\bottomrule
\end{tabular}
\caption{\textbf{Qwen2.5-VL-3B} (direct answers, greedy decoding). Accuracy (\%) on V$^*$ and TreeBench; both RL rows use the same outcome-reward RL.}
\label{tab:qwen25}
\end{table}

\paragraph{Second backbone.}
An adapted direct-answer implementation carries the regional-supervision principle to Qwen2.5-VL-3B (Table~\ref{tab:qwen25}; Appendix~\ref{app:impl}).
Supervised at the query that predicts the answer, alignment SFT already beats CE SFT on V* and TreeBench, and after the same RL stage the aligned pipeline leads by \gainQwen{} points (Table~\ref{tab:stats}).
\section{Analysis}
\label{sec:analysis}

\begin{table}[t]
\centering
\small
\setlength{\tabcolsep}{2pt}
\begin{tabular*}{\columnwidth}{@{\extracolsep{\fill}}lccccc@{}}
\toprule
\multicolumn{6}{@{}l}{\textit{(a) Gradients at the start of RL}} \\
 & $\lVert g\rVert$ & $\cos(-g,\nabla s)$ & \multicolumn{2}{c}{mean $\Delta s$} & $\Delta s{>}0$ \\
\cmidrule(lr){4-5}
 & mixed & mixed & correct & mixed & groups \\
\midrule
Outcome PG & 1.06 & $-$0.04--0.04 & 0 & $-$0.09 & 2/8 \\
KL & 0.11 & $-$0.01--0.05 & $-$0.03 & 0.01 & 3/8 \\
\textbf{Attention} & 0.06 & 0.10--0.32 & 0.18 & 0.13 & 8/8 \\
\bottomrule
\end{tabular*}

\vspace{4pt}
\begin{tabular*}{\columnwidth}{@{\extracolsep{\fill}}lccccc@{}}
\toprule
\multicolumn{6}{@{}l}{\textit{(b) Evidence preference during RL}} \\
Step & 0 & 25 & 50 & held-out & V$^*$ \\
\midrule
$\beta{=}0$ & .760 & .760 & .761 & .770 & .842 \\
\ours{} & .760 & .820 & .860 & .877 & .895 \\
\bottomrule
\end{tabular*}
\caption{\textbf{The attention term supplies the routing signal that outcome RL lacks} (Qwen3.5-4B, same alignment-SFT start). (a) Gradients on rollout groups of five: median norm and cosine with the routing direction $\nabla s$ in mixed groups, mean first-order change of $s$ in all-correct and mixed groups, and how many of these eight groups it raises. (b) $p(G)/(p(G){+}p(B))$ of the supervised heads on 82 fixed training-domain trajectories; held-out: the 17 not used in SFT; V$^*$: at the object words of the base model's V$^*$ reasoning, against the benchmark's own boxes (.844 at the start). Last two columns at step 50.}
\label{tab:rl-mechanism}
\end{table}

\paragraph{Attention RL supplies a direct routing gradient.}
At the alignment-SFT checkpoint, we decompose the RL gradient on the query and key projections of layers 15, 19 and 23 and measure how each objective moves the routing score $s$ of the supervised heads at fact tokens, on 12 rollout groups and eight held-out correct trajectories (Table~\ref{tab:rl-mechanism}a; Appendix~\ref{app:rl-gradients}).
In mixed groups the outcome gradient is more than an order of magnitude larger than the attention term, yet nearly orthogonal to the routing direction, and its effect on $s$ flips sign from group to group.
By contrast, the attention term is small but consistent: it raises $s$ in all eight groups where it is active (cosine 0.10--0.32), including all-correct groups, where every outcome advantage is zero and outcome RL provides no signal at all.
Outcome credit decides which trajectory to reinforce, while the attention term decides where to look; because it always pushes the same way, its effect accumulates.

\paragraph{Routing improves over training.}
We teacher-force the same 82 correct training-domain trajectories through every checkpoint and track the supervised heads (Table~\ref{tab:rl-mechanism}b).
With the attention term, their evidence preference rises from \routStart{} to \routLHA{} in \routStep{} steps, including on the 17 trajectories held out from SFT; without it, the $\beta{=}0$ run stays at \routBzero{} (Appendix~\ref{app:rl-routing}).
The sharper routing carries over to a benchmark that no stage trains on: at the \vstarQueries{} object words in the base model's V* reasoning, the supervised heads prefer V*'s own evidence boxes with \vstarLHA{} after attention RL, against \vstarBzero{} for the control (\vstarStart{} at the start), and their in-box attention share rises from \vstarMassStart{} to \vstarMassLHA{}.

\paragraph{The redirection is visible on a second backbone.}
On Qwen2.5-VL-3B, alignment SFT nearly triples the visual attention inside the V* evidence box at the answer, whereas CE SFT barely moves it, and the change survives RL (Appendix~\ref{app:impl}).

\paragraph{Looking at the right place makes training and inference cheaper.}
Alignment SFT is light: it updates 2.6\% of the language model, and the alignment gradient reaches only the queries of ten heads.
In RL, the aligned policy writes 19\% shorter rollouts than the CE-initialized one, so each step runs 15\% faster; though it recomputes attention, \ours{} runs 8\% faster per step than its $\beta{=}0$ control, as its rollouts are 15\% shorter (Table~\ref{tab:train-cost}).
At test time, alignment SFT$\rightarrow$RL generates \tokSFT\% fewer tokens than CE SFT$\rightarrow$RL and \ours{} \tokBase\% fewer than the base model, not only in the truncated tail (Appendix~\ref{app:efficiency}).

\section{Conclusion}
\label{sec:conclusion}

A thinking VLM usually has the evidence it needs and heads that can find it, but it lacks a training signal that tells those heads where to look.
\ours{} provides this signal by supervising their attention directly.
Grounding boxes and ablations identify the heads, box annotations pick the fact tokens, and one normalizer-free contrast supervises them in SFT and again in RL.
Alignment makes SFT more accurate and the gain survives outcome RL; in RL the attention term keeps sharpening where the heads look.
In a single pass, the resulting 4B model beats inference-time zooming and larger released models with shorter reasoning and faster RL training, and the principle carries over to a direct-answer model.
Localization heads are a practical handle for training perception: boxes are needed only in training, and at test time the model finds the evidence itself.

\section*{Limitations}
We train two Qwen backbones at the 3--4B scale and evaluate on three high-resolution benchmarks of multiple-choice questions; larger models, other model families and open-ended tasks such as charts, documents and video remain to be explored.
\ours{} relies on box annotations for its training questions and on a small set of grounding questions to locate the heads of each new model, so extending it to domains without region annotations would call for automatic grounding.
Fact tokens are identified by matching annotated phrases and counts, which leaves facts stated in other words unsupervised.
Because RL training is expensive, each RL configuration is trained once, and further training seeds would better characterize run-to-run variance.
Finally, the loss reads the attention of the supervised heads during training, which adds training cost, although inference is unchanged.

\bibliography{refs,diagnostics}

\appendix
\renewcommand{\dblfloatpagefraction}{0.75}
\renewcommand{\floatpagefraction}{0.75}
\renewcommand{\textfraction}{0.07}
\renewcommand{\topfraction}{0.9}
\renewcommand{\dbltopfraction}{0.95}
\renewcommand{\bottomfraction}{0.6}
\clearpage
\section*{Appendix}
\begin{itemize}[leftmargin=*,itemsep=0pt,topsep=2pt]
  \item[\ref{app:related}] Extended related work.
  \item[\ref{app:protocol}] Evaluation protocol, scoring audit and paired statistics.
  \item[\ref{app:diag}] Diagnostics behind \S\ref{sec:diagnosis}, including head selection.
  \item[\ref{app:thesis}] Localization heads in Qwen2.5-VL-3B.
  \item[\ref{app:training}] Training details, routing during RL, the RL gradient analysis and the Qwen2.5-VL-3B implementation.
  \item[\ref{app:results}] Greedy decoding, gains by evidence size, response length and inference cost.
  \item[\ref{app:endpoints}] All evaluated endpoints, at temperature 1 and with greedy decoding.
  \item[\ref{app:figures}] Full-width figures.
\end{itemize}

\section{Extended Related Work}
\label{app:related}

\paragraph{Perception in thinking VLMs.}
Longer chains drift toward language priors~\citep{liu2025morethinking}, and the visual re-examination that thinking models verbalize rarely happens~\citep{visualswap}.
Fading attention to the image motivates re-injecting visual information~\citep{sun-etal-2025-mitigating-visual,memvr} or rewarding sustained attention~\citep{remember-r1}.
We locate the problem in a few localization heads.
We supervise where they look inside the image, evidence against background, rather than how much attention the image receives.

\paragraph{Inference-time zooming and tools.}
Guided search~\citep{vstar,zoomeye}, attention-selected crops~\citep{vicrop}, contrastive decoding~\citep{vcd} and crop tools learned with RL~\citep{deepeyes,zhang2025adaptivecof} recover small details at extra inference cost.
\ours{} keeps a single pass and combines with these methods.

\paragraph{RL for perception and grounded reasoning.}
Rule-based RL has been adapted to perception~\citep{yu2025perceptionr1,vision-r1,noisyrollout}.
Grounded-reasoning methods reward boxes written in the chain~\citep{fan2025grit,sarch2025vigorl,treebench} or synthesize harder training data~\citep{zwz}.
Perception-aware objectives act on the policy's outputs.
PAPO adds a KL term between predictions with intact and masked images~\citep{wang2025papo}, VPPO reweights advantages by visual dependency~\citep{huang2025vppo}, and MoCA rewards stated perception through a text-only verifier~\citep{wang2026moca}.
\ours{} supervises the internal routing of a few heads directly, on the fact tokens of correct rollouts.

\paragraph{Supervising attention.}
Attention has long been supervised with word alignments~\citep{liu-etal-2016-neural}, with syntactic parents for one head~\citep{strubell-etal-2018-linguistically} and with human attention in VQA~\citep{qiao2018han,selvaraju2019hint}.
For VLMs, KLAL aligns answer-token attention with boxes~\citep{esmaeilkhani2025klal}.
EASE adds a KL term toward a smoothed evidence prior on the head-averaged attention of one layer, during RL on correct rollouts~\citep{ease}.
Faithful-MR1 supervises the attention of an inserted focus token on a prescribed set of layer--head pairs, followed by RL with a counterfactual attention reward~\citep{tian2026faithfulmr1}.
AVAR combines an attention-guided cold start with visual-attention rewards~\citep{luo2026avar}.
\ours{} differs from all three in its supervision interface.
It targets functional heads selected independently of the training objective, at the native positions that predict visual facts in the model's own reasoning.
Its regional auxiliary loss contrasts the evidence with an equally large background.
This ratio is unaffected by total visual mass and attention sinks~\citep{xiao2024streamingllm,kang2025seewhat}, and the same loss serves SFT and RL.

\paragraph{Attention heads as mechanisms.}
Sparse functional heads are documented in language models~\citep{olsson2022induction,retrieval-heads,zhang-etal-2025-query-focused,chuang-etal-2024-lookback} and in VLMs~\citep{kang2025heads,gandikota2026gazeheads}.
We make them the target of a training loss.

\section{Evaluation Protocol}
\label{app:protocol}
\paragraph{Suite and decoding.}
The suite contains 1,396 questions at native resolution: 191 from V*, 800 from HR-Bench 4K and 405 from TreeBench; a 4K image yields about 8,100 visual tokens.
Qwen3.5-4B models run in thinking mode at temperature 1 with a 32K response budget, served with SGLang, and released models use their own generation format.
Qwen2.5-VL-3B answers directly with greedy decoding (Appendix~\ref{app:impl}); its RL comparison covers the 596 V* and TreeBench questions.
Two decodes of the same checkpoint differ by about 15 correct answers (1.1 points), so trained Qwen3.5-4B models are reported as two-decode means wherever both decodes are available; single-decode rows are marked.

\paragraph{Scoring and audit.}
Rules extract the committed option, and an independent judge model (Qwen3.6-35B-A3B) reads only the responses the rules cannot parse; all Qwen3.5-4B and released models share the rules and the judge.
We checked the scoring of all reported evaluations by hand, including every judge decision, and the automatic scores agreed with the manual labels in every case.
The base-model reference was evaluated once under the frozen protocol and never re-scored.

\paragraph{Separation of training and evaluation data.}
All training questions come from TallyQA, PixMo-Count, Spatial-MQA~\citep{spatialmqa} and VSR.
Before training we removed every question whose image matches an evaluation image, exactly or as a near-duplicate under a perceptual hash (53 questions).
The localization heads are ranked on 96 grounding questions from the RL training pool whose images appear in no SFT trajectory or evaluation set (Appendix~\ref{app:heads}).

\paragraph{Paired comparisons.}
Table~\ref{tab:stats} gives paired bootstrap intervals for the controlled comparisons of \S\ref{sec:experiments}.
Each pair shares questions and decoding seeds; resampling is stratified by benchmark, and HR-Bench's four option rotations are resampled together by source question.

\begin{table}[t]
\centering
\small
\setlength{\tabcolsep}{5pt}
\begin{tabular}{@{}lrcr@{}}
\toprule
Decoding & $\Delta$ & 95\% CI & $+$/$-$ \\
\midrule
\multicolumn{4}{@{}l}{\textit{Alignment SFT$\rightarrow$RL vs.\ CE SFT$\rightarrow$RL}} \\
\quad T=1, decode 1 & $+1.3$ & [$-$0.4, 3.0] & 91/73 \\
\quad T=1, decode 2 & $+2.2$ & [0.5, 3.9] & 92/61 \\
\quad T=1, both decodes & $+1.8$ & [0.5, 2.9] & -- \\
\quad greedy & $+2.1$ & [0.4, 4.0] & 89/59 \\
\midrule
\multicolumn{4}{@{}l}{\textit{\ours{} vs.\ its $\beta{=}0$ control}} \\
\quad T=1, decode 1 & $+0.9$ & [$-$0.9, 2.8] & 84/71 \\
\quad T=1, decode 2 & $+0.9$ & [$-$1.1, 2.7] & 87/75 \\
\quad T=1, both decodes & $+0.9$ & [$-$0.5, 2.2] & -- \\
\midrule
\multicolumn{4}{@{}l}{\textit{Qwen2.5-VL-3B: alignment vs.\ CE SFT$\rightarrow$RL}} \\
\quad greedy, V$^*$+Tree & $+2.5$ & [0.3, 4.7] & 30/15 \\
\bottomrule
\end{tabular}
\caption{\textbf{Paired comparisons between runs that differ in one component.} Accuracy difference (points), 95\% bootstrap interval clustered by source question, and questions gained ($+$) and lost ($-$).}
\label{tab:stats}
\end{table}

\section{Diagnostics}
\label{app:diag}
All numbers in this section are for the frozen Qwen3.5-4B base model under the protocol of Appendix~\ref{app:protocol}.
The exceptions are Appendix~\ref{app:diag-tokens} (Qwen2.5-VL-3B) and the direct-answer models of Appendix~\ref{app:diag-routing}.

\subsection{Evidence masking}
\label{app:diag-mask}
Table~\ref{tab:diag-mask} splits the masking result of \S\ref{sec:diag-evidence} by benchmark and by the area of the evidence box.
The masked and control images share one PNG pipeline, so the evidence-versus-control contrast isolates the evidence.
Masking the control region changes accuracy by $-$1.1 points relative to the original images (CI [$-$4.4, 2.3]).
The drop is largest on V*, whose questions concern a single small object, and it grows as the box shrinks.
Without the evidence, the model also searches longer (9,251 vs.\ 5,784 mean tokens).

\begin{table}[t]
  \centering
  \small
  \setlength{\tabcolsep}{3.5pt}
  \begin{tabular}{lrccc}
    \toprule
    Subset & $n$ & Evid. & Ctrl. & $\Delta$ [95\% CI] \\
    \midrule
    V* & 191 & 43.5 & 82.2 & $-$38.7 [$-$46.6, $-$30.4] \\
    TreeBench & 333 & 33.3 & 45.9 & $-$12.6 [$-$18.6, $-$6.6] \\
    \midrule
    Area $<$0.21\% & 174 & 42.5 & 70.1 & $-$27.6 [$-$36.2, $-$19.0] \\
    0.21--2.4\% & 175 & 37.1 & 62.3 & $-$25.1 [$-$33.7, $-$16.6] \\
    Area $>$2.4\% & 175 & 31.4 & 45.1 & $-$13.7 [$-$22.9, $-$4.6] \\
    \midrule
    All & 524 & 37.0 & 59.2 & $-$22.1 [$-$27.1, $-$17.2] \\
    \bottomrule
  \end{tabular}
  \caption{\textbf{Evidence masking by subset.} Accuracy (\%) with the evidence or an equal-area control region masked; area is the box's fraction of the image.}
  \label{tab:diag-mask}
\end{table}

\subsection{Encoder-side token controls}
\label{app:diag-tokens}
\begin{figure}[!h]
  \centering
  \includegraphics[width=\textwidth]{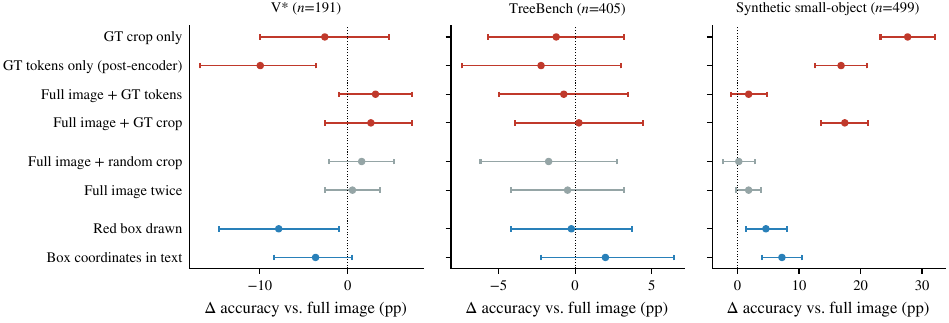}
  \caption{\textbf{Encoder-side token controls on Qwen2.5-VL-3B} (Appendix~\ref{app:diag-tokens}). Accuracy change relative to the full image, with 95\% CIs. Red: evidence given as pixels or post-encoder tokens; grey: input-size and position controls; blue: location cues.}
  \label{fig:diag-tokens}
\end{figure}

A matrix of input interventions on Qwen2.5-VL-3B separates what the vision encoder preserves from what the language model selects (Figure~\ref{fig:diag-tokens}).
The runs use native resolution (up to 12.8M pixels), greedy decoding, a 4,096-token budget and one LLM judge.
We report paired changes against the full image.
The key condition gives the language model only the post-encoder visual tokens inside the ground-truth box, taken from the encoding of the full image.
These tokens are a subset of those the model already receives.
We remove the others to locate where the evidence is lost, not for efficiency as in token pruning~\citep{chen2024fastv,zhang2025sparsevlm}.
Two conditions control for input size and position: the full image given twice, and the full image plus a random crop.

On 499 rendered small-object scenes, keeping only the box tokens raises accuracy by 16.8 points (CI [12.6, 21.0]), and a re-encoded crop of the box raises it by 27.7 (CI [23.2, 32.1]).
Appending the same box tokens to the full image adds 1.8 points (CI [$-$1.0, 4.8]), and the size and position controls add 1.8 and 0.2.
On these synthetic scenes the evidence therefore survives encoding, and removing the other tokens helps more than repeating the evidence.
\paragraph{Qwen3.5-4B on natural images.}
We repeat the token edits on Qwen3.5-4B in thinking mode for the 524 V* and TreeBench questions whose boxes admit an equal-area control (Table~\ref{tab:token-controls-q35}).
The edits remove visual tokens after the vision encoder has processed the full image and keep the remaining tokens at their original positions.
Keeping only the tokens inside the evidence box preserves accuracy (60.7\% against 60.1\% with the full image), whereas keeping an equal-area control set instead drops it to 38.0\% (box minus control: $+$22.7 points, CI [17.7, 27.9]).
Appending the box tokens to the full image changes accuracy by only $+$0.8 points.
The box tokens alone thus carry the answer, and giving the model more evidence does not help; selecting it does.
On V*, keeping only the box tokens costs some context ($-$6.3 points); on TreeBench, whose questions relate several boxed objects, accuracy is even slightly higher ($+$4.5).
For Qwen2.5-VL-3B on V*, box tokens alone cost 9.9 points (CI [$-$16.8, $-$3.7]).
Because answers draw on some context around the box, \ours{} contrasts the evidence with an equally large background inside the image instead of removing the rest of it.

\begin{table}[t]
  \centering
  \small
  \setlength{\tabcolsep}{3.5pt}
  \begin{tabular}{lccc}
    \toprule
    Visual tokens given & V* & Tree & All \\
    \midrule
    Full image & 83.8 & 46.5 & 60.1 \\
    Evidence box only & 77.5 & 51.1 & 60.7 \\
    Equal-area control only & 44.5 & 34.2 & 38.0 \\
    Full image + box tokens & 83.8 & 47.7 & 60.9 \\
    Full image + control tokens & 80.6 & 45.6 & 58.4 \\
    \bottomrule
  \end{tabular}
  \caption{\textbf{Post-encoder token controls on Qwen3.5-4B} (thinking mode, one decode; 191 V* and 333 TreeBench questions). Accuracy (\%).}
  \label{tab:token-controls-q35}
\end{table}

\subsection{Localization head selection}
\label{app:heads}
We score each of the 128 heads of the eight full-attention layers on 96 held-out grounding questions.
They are drawn from the RL training pool and share no images with the SFT trajectories or any evaluation set.
For a head, the score is $\log(p_G+\epsilon) - \log(\bar p_C+\epsilon)$.
Here $p_G$ is the share of its attention over visual tokens that falls in the evidence box, and $\bar p_C$ is the mean share over two equal-area control boxes.
Accuracy is never used for selection.
The four highest-scoring heads are L19H14, L15H11, L19H2 and L23H9.
The ranking computed independently on V* correlates with this ranking (Spearman 0.80) and shares six of the top eight heads.
\ours{} supervises the top ten (three in layer 15, six in layer 19, one in layer 23).

\subsection{Localization-head ablation}
\label{app:diag-ablation}
Table~\ref{tab:diag-ablation} gives the generation ablation of \S\ref{sec:diag-heads} per benchmark.
The drop appears on all three benchmarks.

\begin{table}[t]
  \centering
  \small
  \begin{tabular}{lcccc}
    \toprule
    & Base & R4 & L4 & L4$-$R4 [95\% CI] \\
    \midrule
    V* & 84.8 & 83.2 & 69.1 & $-$14.1 [$-$19.9, $-$8.4] \\
    HR-4K & 84.0 & 83.3 & 68.3 & $-$15.0 [$-$18.0, $-$12.0] \\
    Tree & 49.6 & 49.6 & 40.5 & $-$9.1 [$-$13.8, $-$4.4] \\
    \midrule
    All & 74.1 & 73.5 & 60.3 & $-$13.2 [$-$15.5, $-$10.9] \\
    \bottomrule
  \end{tabular}
  \caption{\textbf{Localization-head ablation by benchmark.} Accuracy (\%) of the base model and with four random (R4) or localization (L4) heads ablated.}
  \label{tab:diag-ablation}
\end{table}

A forced-choice probe isolates the answer decision from generation.
We select the top eight heads on 96 V* questions and test on the other 95, reading the option distribution directly after an empty thinking block.
On the held-out questions, the selected heads place more attention on the answer box than layer-matched random heads ($+$0.95 in log ratio, CI [0.81, 1.10]).
Zeroing them lowers forced-choice accuracy from 52.6\% to 36.8\%.
Layer-matched random or bottom-ranked heads leave it at 52.6\% and 51.6\%.

\subsection{Where the image is read: direct answers and thinking}
\label{app:diag-routing}
\paragraph{Direct answers.}
Correct direct answers have a higher evidence share than incorrect ones across six models.
The models are Qwen2.5-VL-3B and 7B~\citep{qwen25vl}, LLaVA-OneVision-7B~\citep{llavaov}, InternVL3-8B~\citep{internvl3}, Pixel-Reasoner~\citep{pixelreasoner} and a region-supervised Qwen2.5-VL-3B.
For the five top-ranked localization heads of each model, we measure the share of their visual attention inside the evidence box at the first answer token.
Across 12 model--dataset settings on V*, TreeBench and a held-out set, correct answers have the higher share in all 12, by 0.040 to 0.113, with CIs above zero in nine.

\begin{figure}[t]
  \centering
  \includegraphics[width=0.62\columnwidth]{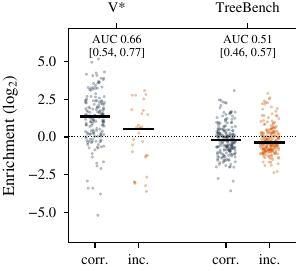}
  \caption{\textbf{Question-time evidence enrichment of the four localization heads} for correct and incorrect base answers (log scale; bars: medians; dotted: uniform attention).}
  \label{fig:diag-routing}
\end{figure}

\paragraph{Qwen3.5-4B before reasoning.}
Figure~\ref{fig:diag-routing} plots the evidence enrichment of the four localization heads at the last question token, for questions the base model answers correctly and incorrectly.
Enrichment is the share of their visual attention inside the ground-truth box divided by the box's share of visual tokens.
For boxes smaller than one visual token we use the token that contains the box center.
On V*, enrichment exceeds 1 for 80\% of correct and 55\% of incorrect answers (medians 2.56 and 1.44).
It predicts correctness with AUC 0.66 (CI [0.54, 0.77]; 162 correct, 29 incorrect).
TreeBench questions relate several objects that the reasoning visits one by one.
Their evidence is read at the fact tokens of the chain rather than at the question end, which is where \ours{} places its supervision.

\paragraph{Qwen3.5-4B along the reasoning.}
We take 200 training questions outside the head-selection images, 50 from each of the four source datasets, that the frozen base model answers correctly in one to four of five samples.
For each question we keep two finished rollouts, one correct and one incorrect, each under 8,000 tokens.
We teacher-force these rollouts and record the ten supervised heads at fact tokens in the first and second half of the chain.
The share is $p(G)/(p(G){+}p(B))$, averaged over queries, heads and rollouts, with $G$ the box of the stated object and $B$ a disjoint equal-area background.
It is 0.77 in the first half of the chain and 0.72 in the second (median enrichment 3--3.6).
At the token that states the final answer it is 0.50, the value of uniform attention.
The image is thus read where facts are stated, and this reading weakens as the chain grows.

\paragraph{Qwen3.5-4B on benchmark rollouts.}
We repeat the measurement on 220 base rollouts of 110 V* and TreeBench questions, one correct and one incorrect per question, at eight positions sampled from each tenth of the reasoning.
The evidence set is the union of the question's annotated boxes.
The share falls steadily from 0.71 in the first tenth of the chain to 0.61 in the last, and to 0.59 at the answer, in correct and incorrect rollouts alike.

\subsection{Decomposing the drift of visual attention}
\label{app:diag-decay}
\begin{figure}[!h]
  \centering
  \includegraphics[width=\textwidth]{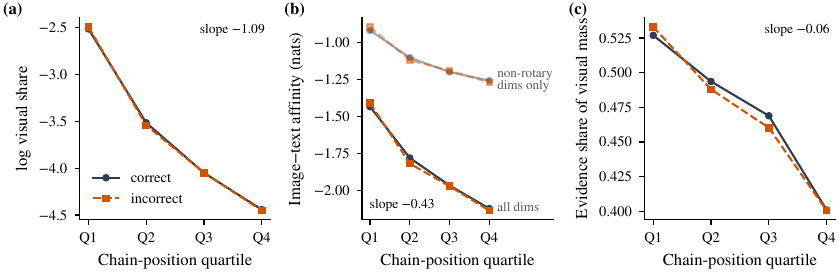}
  \caption{\textbf{Visual attention drifts away from the evidence as the chain grows} (Appendix~\ref{app:diag-decay}; four localization heads, 1,072 same-question rollout pairs). (a) Log share of attention on visual tokens. (b) Image--text logit gap $\Delta$, from all and from non-rotary dimensions. (c) Share of visual attention on the evidence box.}
  \label{fig:diag-decay}
\end{figure}

The drift of visual attention splits into dilution by the growing text and a weakening preference for the image.
For a head and query position, let $m_{\text{vis}}$ be its attention mass on image keys, and let $n_{\text{img}}$ and $n_{\text{text}}$ be the numbers of image and text keys (excluding the first four sink tokens).
Let $\Delta = \operatorname{LME}(z_{\text{img}}) - \operatorname{LME}(z_{\text{text}})$ be the gap between the log-mean-exp logits of the two key sets.
Then, ignoring sinks,
\begin{equation*}
  \log m_{\text{vis}} = \log \sigma\bigl(\Delta + \log n_{\text{img}} - \log n_{\text{text}}\bigr),
\end{equation*}
so the decay splits into a dilution term, $\log n_{\text{text}}$, and an affinity term, $\Delta$.
We teacher-force 1,072 same-question pairs of best-of-five base rollouts on training questions (2,144 rollouts, median thinking length about 3,300 tokens).
We record the four localization heads along the chain (Figure~\ref{fig:diag-decay}).
The log visual share falls with slope $-$1.09 against $\log n_{\text{text}}$, close to pure dilution by the growing text.
The affinity $\Delta$ falls as well (slope $-$0.43, CI [$-$0.46, $-$0.40]).
It still falls at $-$0.24 (CI [$-$0.26, $-$0.21]) when computed from the non-rotary query and key dimensions alone.
The heads' content preference for the image therefore weakens, in addition to their positional reach.
Within the image, on the 1,040 pairs (2,080 rollouts) with valid evidence boxes, the share on the evidence box drops from 0.53 to 0.40 between the first and last quarter of the chain (slope $-$0.06 against $\log n_{\text{text}}$, Figure~\ref{fig:diag-decay}c).
Over all 128 full-attention heads the affinity slope is $-$0.15, so the localization heads drift faster than the average head.

Table~\ref{tab:diag-bias} adds $\beta \log(n_{\text{text}}/n_0)$ to the image-key logits of the four heads at every decoding step, where $n_0$ is the number of text keys at the end of the prompt.
At $\beta{=}1$ this bias offsets the dilution term.
Accuracy stays at the base level.
Restoring how much attention reaches the image does not tell the heads where in it to look, and that is what the contrastive loss of \ours{} supervises.

\begin{table}[t]
  \centering
  \small
  \setlength{\tabcolsep}{4pt}
  \begin{tabular}{lcccc}
    \toprule
    $\beta$ & Acc. & $\Delta$ vs.\ base [95\% CI] & Trunc. & Tokens \\
    \midrule
    0 (base) & 74.1 & -- & 37 & 4,678 \\
    0.25 & 74.1 & $-$0.1 [$-$1.9, 1.7] & 48 & 4,751 \\
    0.5 & 73.9 & $-$0.3 [$-$2.0, 1.4] & 42 & 5,013 \\
    1 & 74.2 & $+$0.1 [$-$1.8, 2.0] & 54 & 5,100 \\
    \bottomrule
  \end{tabular}
  \caption{\textbf{Length-aware visual logit bias at the four localization heads} (1,396 questions, one decode). Trunc.: budget-truncated responses; Tokens: mean completion tokens.}
  \label{tab:diag-bias}
\end{table}

\section{Localization Heads in Qwen2.5-VL-3B}
\label{app:thesis}
This section analyzes Qwen2.5-VL-3B-Instruct answering directly, without thinking.
Attention is read at the query that produces the answer.
The evidence share of a head is the fraction of its visual attention that falls on the visual tokens inside the ground-truth box.
Layers and heads are indexed from 0.

\begin{figure}[t]
  \centering
  \includegraphics[width=\columnwidth]{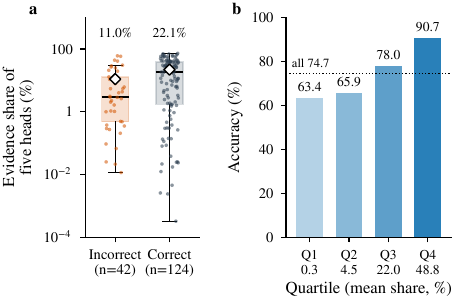}
  \caption{\textbf{Evidence routing predicts answer correctness} (Qwen2.5-VL-3B, 166 V* questions, five fixed localization heads at the answer position). (a) Evidence share for correct and incorrect greedy answers; diamonds: means. (b) Accuracy by quartile of the share.}
  \label{fig:thesis-routing}
\end{figure}

\begin{figure}[!h]
  \centering
  \includegraphics[width=\textwidth]{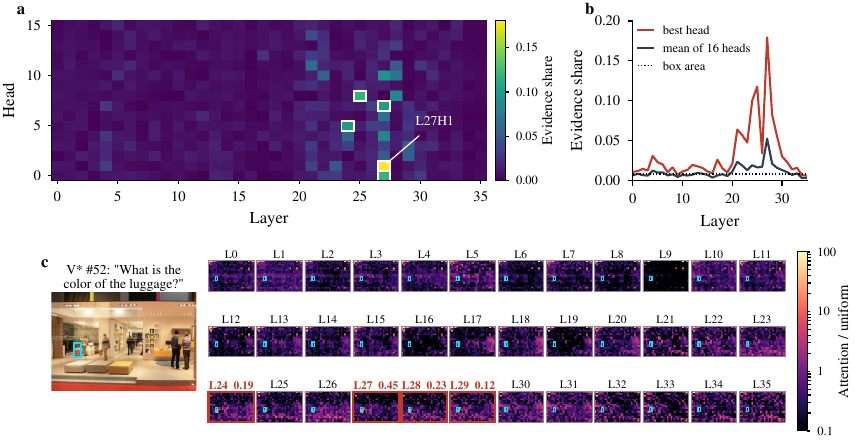}
  \caption{\textbf{A few deep heads localize the evidence} (Appendix~\ref{app:thesis}; Qwen2.5-VL-3B, direct answers). (a) Evidence share of each of the 576 heads on V*; white boxes mark the five strongest. (b) Strongest head and head mean per layer; dotted: mean box area. (c) Attention of each layer on one V* question (cyan box), relative to uniform; red layers have evidence share $\geq$0.10.}
  \label{fig:thesis-heads}
\end{figure}

\paragraph{Localization heads are sparse and deep.}
Over the 191 V* questions, 17 of the 576 heads place more than 5\% of their visual attention on the evidence box, which covers 0.83\% of the visual tokens on average (Figure~\ref{fig:thesis-heads}a).
The strongest head, L27H1, places 17.9\% there, and the five strongest lie in layers 24 to 27.
On a single question the same pattern appears layer by layer, as the evidence share stays at or below 0.03 through layer 22 and peaks at 0.45 in layer 27 (Figure~\ref{fig:thesis-heads}c).

\paragraph{Routing predicts correctness.}
On 166 V* questions, five fixed localization heads place 22.1\% of their visual attention on the evidence when the answer is correct and 11.0\% when it is wrong (Mann--Whitney $p{=}0.006$; Figure~\ref{fig:thesis-routing}).
Accuracy rises from 63.4\% in the lowest quartile of this share to 90.7\% in the highest (point-biserial $r{=}0.24$).

\begin{figure}[!h]
  \centering
  \includegraphics[width=0.87\textwidth]{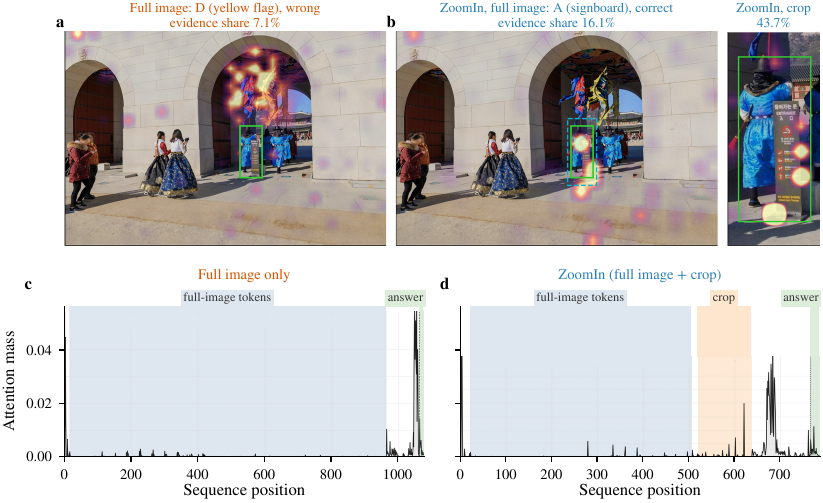}
  \caption{\textbf{TreeBench \#181: the crop pulls attention onto the occluding signboard} (Qwen2.5-VL-3B). (a) From the full image the model answers D, the yellow dragon flag, and its localization heads put 7.1\% of their visual attention on the evidence (green box). (b) With the crop appended it answers A, the signboard, and the share rises to 59.7\%. (c, d) Attention mass of the same heads over the input sequence.}
  \label{fig:thesis-case-tree}
  \vspace{0.6em}
  \includegraphics[width=0.87\textwidth]{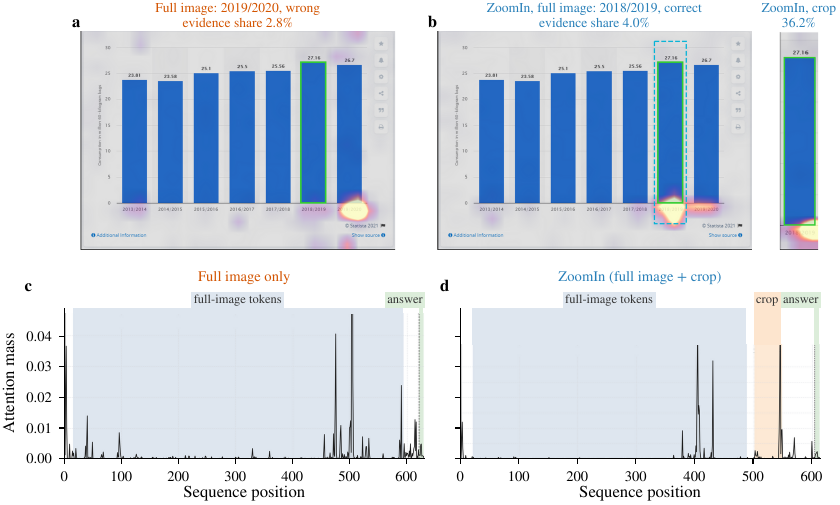}
  \caption{\textbf{RefChartQA \#207: the crop isolates the maximal bar.} (a) Full image: wrong answer 2019/2020, evidence share 2.8\%. (b) With the crop: correct answer 2018/2019, 40.3\%. (c, d) As in Figure~\ref{fig:thesis-case-tree}.}
  \label{fig:thesis-case-chart}
\end{figure}

\paragraph{Adding a crop moves attention onto the evidence.}
Figures~\ref{fig:thesis-case-tree} and~\ref{fig:thesis-case-chart} show two questions that the model answers wrongly from the full image.
It answers both correctly once a crop around the evidence box is appended as a second image (ZoomIn).
The evidence share of eight localization heads at the answer tokens rises from 7.1\% to 59.7\% on TreeBench and from 2.8\% to 40.3\% on RefChartQA.
Most of the added share falls on the crop.

\section{Training Details}
\label{app:training}
\subsection{Alignment SFT on Qwen3.5-4B}
We train the query, key, value and output projections of the full-attention layers 15, 19 and 23 (12 matrices, 110M parameters).
Training uses FP32 with AdamW, a constant learning rate of $10^{-5}$, batch size 16 and gradient clipping at 1.
Alignment is data-efficient.
A compact set of 65 correct best-of-five responses of the base model suffices, drawn from the box-annotated training questions of \S\ref{sec:align} (34 PixMo-Count, 29 TallyQA and 2 spatial questions) and disjoint from the head-selection set; a fixed hash of each question splits a pool of 82 into these 65 and 17 held-out trajectories used for routing measurements (Appendix~\ref{app:rl-routing}).
The alignment loss acts on the predictors of 44 fact tokens, in the 15 trajectories whose reasoning names an annotated object or states its count exactly as annotated; a phrase is eligible only if it does not occur in the question and its context passes a manual review, and coordinates generated by the model are never used as evidence.
This compact set of supervised positions already yields the gains of Table~\ref{tab:main}.
Fact-based selection concentrates the supervision: of 64 query positions sampled uniformly from each trajectory, 13.7\% fall on fact content, against 98.6\% for fact-first selection.
Training runs for 32 updates.
The alignment term is computed with hidden states and keys detached.
Its background set $B_i$ is fixed before training: the visual tokens outside every annotated box that lie nearest to the centroid of $G_i$, as many as in $G_i$.
CE SFT uses exactly the same data order, updates and parameters with $\alpha{=}0$.
Table~\ref{tab:stats} reports the paired evaluation of the matched SFT$\rightarrow$RL pipelines under sampled and greedy decoding.

\subsection{Attention RL on Qwen3.5-4B}
\label{app:rl}
We use verl with GRPO, 16 questions $\times$ 8 rollouts per step, temperature 1, a 32,768-token response budget and a low-variance KL loss of weight 0.01.
All parameters are trained, and attention uses FlashAttention-2.
CE SFT$\rightarrow$RL and alignment SFT$\rightarrow$RL in Table~\ref{tab:main} use learning rate $5\times10^{-7}$ with one update per step and train for 100 steps on the first 1,600 training questions, in the same fixed order.
\ours{} and its $\beta{=}0$ control (Table~\ref{tab:stats}) start from the same alignment-SFT checkpoint and differ only in $\beta$.
They use a constant learning rate of $10^{-6}$ and mini-batches of 8 questions (two optimizer updates per step) on the 3.3K training questions that the base model answers correctly in one to four of five samples, in the same fixed order, and both train for \rlStep{} steps.
The attention branch of the shared training code is enabled only for $\beta{>}0$.
Each run uses four GPUs of the same node.
Aligned policies write shorter rollouts and therefore train faster (Table~\ref{tab:train-cost}); this holds for \ours{} although it recomputes the attention of the supervised heads.

\begin{table}[t]
  \centering
  \small
  \begin{tabular}{lrr}
    \toprule
    Run & s / step & Rollout tokens \\
    \midrule
    \multicolumn{3}{l}{\textit{Outcome RL, 100 steps}} \\
    CE SFT$\rightarrow$RL & 448 & 3,170 \\
    Alignment SFT$\rightarrow$RL & \textbf{379} ($-$15\%) & \textbf{2,570} ($-$19\%) \\
    \midrule
    \multicolumn{3}{l}{\textit{Attention RL, 50 steps}} \\
    $\beta{=}0$ control & 667 & 6,179 \\
    \ours{} & \textbf{613} ($-$8\%) & \textbf{5,230} ($-$15\%) \\
    \bottomrule
  \end{tabular}
  \caption{\textbf{Training cost of RL} (Qwen3.5-4B, four GPUs of one node). Wall-clock time per step and mean rollout length, averaged over training; the two runs of each pair use the same RL data order and hyperparameters.}
  \label{tab:train-cost}
\end{table}

\paragraph{Compute.}
All Qwen3.5-4B training and evaluation runs on one node with eight NVIDIA H20 GPUs (96\,GB each); Qwen2.5-VL-3B runs on a server with eight NVIDIA A100-80GB GPUs.
On the H20 node, alignment SFT and CE SFT each take about 0.8 GPU-hours on one GPU.
CE SFT$\rightarrow$RL and alignment SFT$\rightarrow$RL take about 50 and 42 GPU-hours, \ours{} and its $\beta{=}0$ control about 34 and 37, and the EASE baseline about 46, each on four GPUs.
One decode of the 1,396 evaluation questions takes about two GPU-hours.
Including exploratory runs, the project used the H20 node for about eight weeks, at most about 11K GPU-hours, and the A100 server for about a month, at most about 6K GPU-hours, for Qwen2.5-VL-3B and external baselines.

\paragraph{Attention term.}
The term supervises at most 64 fact positions per correct rollout: object nouns of the annotated phrases and enumerated counts inside the reasoning, excluding restated answers, coordinates and option letters.
It needs the attention weights of the ten supervised heads at these queries, which we recompute explicitly.
For each rollout the evidence set is the union of the annotated boxes it refers to, and the background is an equally large random set of visual tokens disjoint from every box.
Like the policy loss, the term is averaged over the rollouts of each mini-batch; hidden states and keys are not detached, and advantages are left unchanged.

\begin{figure}[t]
  \centering
  \includegraphics[width=\columnwidth]{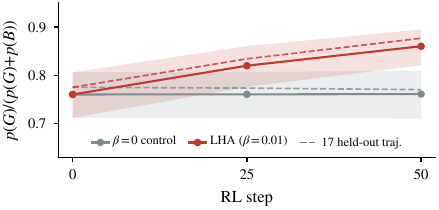}
  \caption{\textbf{Evidence preference along RL} (Qwen3.5-4B, same alignment-SFT start). The ten supervised heads at the fact tokens of 82 fixed training-domain trajectories, teacher-forced through each checkpoint; dashed: the 17 held out from SFT; bands: 95\% bootstrap intervals.}
  \label{fig:rl-routing}
\end{figure}

\paragraph{Routing on fixed trajectories.}
\label{app:rl-routing}
We teacher-force 82 fixed trajectories, the 65 alignment trajectories and 17 held-out ones (135 fact positions), through each RL checkpoint and measure $p(G)/(p(G){+}p(B))$ of the ten supervised heads at the fact positions (Figure~\ref{fig:rl-routing}).
It is \routStart{} at the alignment-SFT start; after \routStep{} steps it is \routLHA{} with the attention term and \routBzero{} without it, a paired difference of $+$\routDiff{} \routDiffCI{} over trajectories, with \routAbove\% of trajectories higher under the attention term.

\subsection{Gradient decomposition of the RL objective}
\label{app:rl-gradients}
\begin{figure}[t]
  \centering
  \includegraphics[width=0.92\columnwidth]{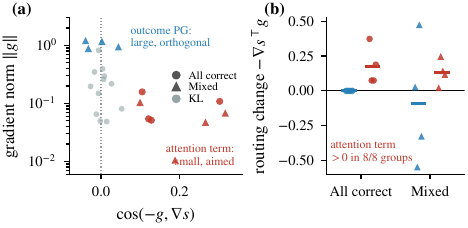}
  \caption{\textbf{Gradient decomposition at the alignment-SFT policy} (Qwen3.5-4B; points: groups of five rollouts). (a) Norm of each objective's gradient against its cosine with the routing direction $\nabla s$. (b) First-order change of $s$ under one gradient step (bars: means); all-wrong groups receive neither term.}
  \label{fig:gradient-decomposition}
\end{figure}
Figure~\ref{fig:gradient-decomposition} measures which part of $\mathcal{L}_{\text{RL}}$ moves the routing of the supervised heads.
We take gradients at the alignment-SFT policy of Qwen3.5-4B with respect to the query and key projections of layers 15, 19 and 23.
The routing score $s=\log(p_G/p_B)$ is the attention of the ten supervised heads on the evidence $G$ relative to an equally large background $B$.
It is measured at the queries that predict fact tokens and averaged over eight held-out correct trajectories.
Each objective's gradient $g$ is computed on 12 groups of five rollouts sampled from the base model: four all-correct, four mixed and four all-wrong.
We use the RL weights, $\beta{=}0.01$ and a KL coefficient of 0.01 toward the base model as reference.
For each objective, $-\nabla s^{\top}g$ is the first-order change of $s$ under a gradient step, and the cosine between $-g$ and $\nabla s$ measures its alignment with the routing direction.
The attention term gives $-\nabla s^{\top}g$ from 0.02 to 0.37 and a cosine from 0.10 to 0.32 in the eight groups where it is active.
In all-correct groups every advantage is zero, so the outcome policy gradient vanishes while the attention term contributes a mean gain of 0.18.
In mixed groups the outcome policy gradient is about 18 times larger than the attention term, with a cosine to $\nabla s$ between $-0.04$ and $0.04$.
The KL term stays small in every group, so using the base model rather than the aligned start as its reference has little effect on the decomposition.

\subsection{Qwen2.5-VL-3B implementation}
\label{app:impl}
\label{app:qwen25}
The Qwen2.5-VL-3B validation applies the same regional-supervision principle in the no-think setting.
It trains an evidence-versus-background loss on localization heads in SFT, follows it with outcome-reward RL, and compares against CE SFT$\rightarrow$RL, which drops only the alignment term.
Its implementation differs from the Qwen3.5-4B algorithm as follows (Qwen3.5-4B first, Qwen2.5-VL-3B second).
\begin{itemize}[leftmargin=*,itemsep=1pt,topsep=2pt]
  \item \textbf{Supervised query}: fact tokens inside the reasoning; the query that predicts the first answer token.
  \item \textbf{Heads}: ten fixed heads ranked on held-out grounding questions (Appendix~\ref{app:heads}); the five heads with the highest localization score on each training example.
  \item \textbf{Background}: an equal-size set outside every box; all non-evidence visual tokens.
  \item \textbf{Trainable parameters}: full attention projections of layers 15, 19 and 23 (2.6\% of the language model), alignment gradient to queries only; LoRA (rank 16) on the attention projections of all 36 layers.
  \item \textbf{Weighting}: adaptive $\alpha$ of Eq.~\eqref{eq:alpha}; a fixed weight of 0.5.
  \item \textbf{SFT data}: 65 self-generated correct trajectories (32 updates); 30K annotated short-answer examples (1,000 updates), with multi-box examples trained by cross-entropy only.
  \item \textbf{RL}: GRPO with outcome reward, 16$\times$8 rollouts (Appendix~\ref{app:rl}); GRPO with an LLM-judged outcome reward plus a 0.2 format score, 32$\times$8 rollouts, 300 steps; all parameters trained in both.
  \item \textbf{Evaluation}: thinking mode at temperature 1 (Appendix~\ref{app:protocol}); direct answers with greedy decoding, scored by rule matching on V*, an LLM judge on TreeBench and the main-text scorer on HR-Bench 4K. The RL comparison of Qwen2.5-VL-3B is evaluated on V* and TreeBench.
\end{itemize}

\begin{figure}[t]
  \centering
  \includegraphics[width=\columnwidth]{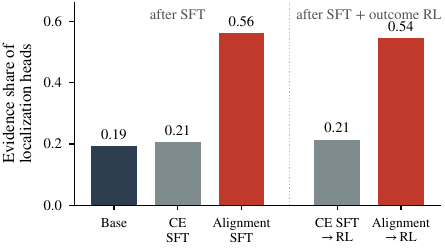}
  \caption{\textbf{Alignment SFT raises the evidence share at the answer position, and RL preserves it} (Qwen2.5-VL-3B, V*). Grey: CE; red: alignment.}
  \label{fig:routing-training}
\end{figure}

\paragraph{Routing after each stage.}
Figure~\ref{fig:routing-training} measures the share of visual attention inside the V* evidence box at the query that predicts the first answer token.
The share is Eq.~\eqref{eq:share} with $R$ the evidence box, averaged over the five heads with the highest share on each question, on 164--166 paired V* questions per model.
These heads are chosen per question with the ground-truth box, as in training.
The measure therefore shows that the training objective is optimized, and it complements the fixed-head diagnostics of \S\ref{sec:diagnosis}.
The mean share is 0.19 for the base model, 0.21 after CE SFT and after CE SFT$\rightarrow$RL, 0.56 after alignment SFT and 0.54 after alignment SFT$\rightarrow$RL.
A question counts as focused when the share is at least 0.3.
The focused fraction is 28.9\% for the base model, 31.1\% after CE SFT and 63.4\% after alignment SFT.
It is 33.7\% after CE SFT$\rightarrow$RL and 64.5\% after alignment SFT$\rightarrow$RL.

\section{Additional Results}
\label{app:results}
\subsection{Inference cost}
\label{app:cost}
\begin{table}[!ht]
\centering
\small
\setlength{\tabcolsep}{4pt}
\begin{tabular}{@{}lrrrr@{}}
\toprule
Method & Passes & +Img. & Tokens & Acc. \\
\midrule
Base\textsuperscript{\dag} & 1 & 0 & 4,678 & 74.1 \\
ViCrop & 3 & 3 & 4,230 & 75.9 \\
VCD\textsuperscript{\dag} & 2$\times$/token & 1 & 4,612 & 75.3 \\
ZoomEye\textsuperscript{\dag} & 124.2 & 123.2 & 5,371 & 72.1 \\
Fixed-grid zoom\textsuperscript{\dag} & 1 & 4 & 4,725 & 74.1 \\
GRPO & 1 & 0 & 4,828 & 74.0 \\
DAPO & 1 & 0 & 4,011 & 73.6 \\
EASE & 1 & 0 & 5,316 & 73.6 \\
\midrule
CE SFT $\rightarrow$ RL & 1 & 0 & 3,876 & 74.4 \\
Alignment SFT $\rightarrow$ RL & 1 & 0 & 3,112 & 76.1 \\
\quad + head-guided zoom & 3 & 4 & 2,972 & 76.6 \\
\textbf{\ours{}} & 1 & 0 & 3,541 & 76.6 \\
\bottomrule
\end{tabular}
\caption{\textbf{Inference cost per question.} Passes: full forward passes over the prompt (VCD: two per decoding step); +Img.: image encodings beyond the original image; Acc.: overall accuracy (\%).}
\label{tab:cost}
\end{table}

Table~\ref{tab:cost} counts passes and extra image encodings.

\subsection{Greedy decoding}
\label{app:t0}
Alignment SFT$\rightarrow$RL is also the most accurate model under greedy decoding (Table~\ref{tab:t0}).
Greedy decoding makes the base model loop: 266 of 1,396 responses (19.1\%) are budget-truncated, compared with 37 at temperature 1, so all accuracies are lower.
Its gain over CE SFT$\rightarrow$RL ($+$2.1 points, CI [0.4, 4.0]) is close to that at temperature 1, and it generates 24.1\% fewer completion tokens.

\subsection{Gains by evidence size}
\label{app:gain-size}
On the 524 V* and TreeBench questions with an equal-area control (Appendix~\ref{app:diag-mask}), we split the questions into tertiles by the fraction of the image covered by the evidence boxes and compare alignment SFT$\rightarrow$RL with CE SFT$\rightarrow$RL, averaging both decodes.
Alignment SFT$\rightarrow$RL gains over CE SFT$\rightarrow$RL in every tertile: $+$2.6 points for the smallest boxes (below 0.25\% of the image), $+$1.1 for the middle tertile and $+$2.0 for the largest boxes (above 2.6\%).
On the 800 HR-Bench 4K questions it gains 1.7 points (CI [0.2, 3.2]).

\subsection{Response length}
\label{app:efficiency}
\begin{figure}[!h]
  \centering
  \includegraphics[width=0.5\textwidth]{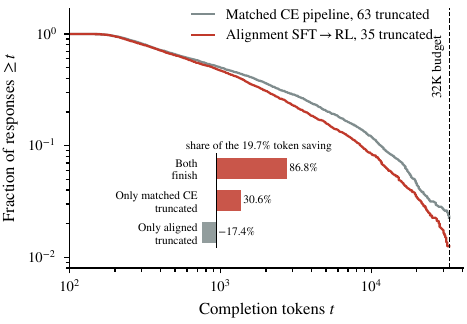}
  \caption{\textbf{The alignment loss shortens the whole length distribution.} Survival curves of response length over two decodes; inset: token saving split by which model finishes; dashed line: 32K budget.}
  \label{fig:efficiency}
\end{figure}

Over both decodes, alignment SFT$\rightarrow$RL is shorter than CE SFT$\rightarrow$RL throughout the length distribution, not only in the truncated tail (Figure~\ref{fig:efficiency}).
It generates 8.69M tokens against 10.82M (19.7\% fewer) and is budget-truncated on 35 instead of 63 responses; in 44 of the 60 pairs where only one model is truncated, the aligned model finishes.
Pairs where both finish carry 86.8\% of the saving.
On V*, HR-Bench 4K and TreeBench it answers 12, 27 and 10 more questions with 23.4\%, 18.0\% and 21.3\% fewer tokens.

\clearpage
\onecolumn
\section{All Endpoints}
\label{app:endpoints}
Table~\ref{tab:appendix-endpoints} lists every thinking-mode evaluation in the paper, with both decodes of each trained model, and Table~\ref{tab:t0} gives the greedy-decoding evaluation of Appendix~\ref{app:t0}.

\footnotesize
\setlength{\tabcolsep}{4pt}
\renewcommand{\arraystretch}{0.75}
\begin{longtable}{@{}lcrrrrrr@{}}
\caption{All thinking-mode endpoints (temperature 1, 32K budget) used in the paper. V$^*$, HR-4K and Tree give correct answers out of 191, 800 and 405; Overall is accuracy in \% with the count out of 1,396; Tokens is the mean completion length and Tr.\ the number of budget-truncated responses. Dec.\ indexes the two independent decodes (1 and 2) of the same checkpoint; HR-Bench rotations of one image are scored as separate questions.}\label{tab:appendix-endpoints}\\
\toprule
Endpoint & Dec. & V$^*$ & HR-4K & Tree & Overall & Tokens & Tr. \\
\midrule
\endfirsthead
\toprule
Endpoint & Dec. & V$^*$ & HR-4K & Tree & Overall & Tokens & Tr. \\
\midrule
\endhead
\midrule
\multicolumn{8}{r}{\textit{continued on next page}} \\
\endfoot
\bottomrule
\endlastfoot
\multicolumn{8}{@{}l}{\textit{Qwen3.5-4B (base)}} \\*
\quad Qwen3.5-4B (base) & 1 & 162 & 672 & 201 & 74.1 (1035) & 4,678 & 37 \\
\midrule
\multicolumn{8}{@{}l}{\textit{Head ablation at inference (base model)}} \\*
\quad Ablate the four localization heads & 1 & 132 & 546 & 164 & 60.3 (842) & 11,471 & 327 \\
\quad Ablate four random heads (same layers) & 1 & 159 & 666 & 201 & 73.5 (1026) & 3,384 & 24 \\
\midrule
\multicolumn{8}{@{}l}{\textit{RL from the base model (1,810-question pool, 100 updates)}} \\*
\quad GRPO & 1 & 163 & 664 & 201 & 73.6 (1028) & 4,875 & 40 \\
\quad GRPO & 2 & 159 & 672 & 208 & 74.4 (1039) & 4,780 & 41 \\
\quad DAPO & 1 & 153 & 678 & 200 & 73.9 (1031) & 3,999 & 27 \\
\quad DAPO & 2 & 160 & 654 & 210 & 73.4 (1024) & 4,023 & 28 \\
\quad EASE & 1 & 158 & 672 & 195 & 73.4 (1025) & 5,156 & 61 \\
\quad EASE & 2 & 163 & 667 & 201 & 73.9 (1031) & 5,476 & 66 \\
\midrule
\multicolumn{8}{@{}l}{\textit{Training on self-generated trajectories}} \\*
\quad CE SFT & 1 & 156 & 678 & 209 & 74.7 (1043) & 3,730 & 28 \\
\quad Alignment SFT & 1 & 164 & 683 & 215 & 76.1 (1062) & 3,940 & 32 \\
\quad CE SFT $\rightarrow$ RL & 1 & 159 & 680 & 208 & 75.0 (1047) & 3,843 & 24 \\
\quad CE SFT $\rightarrow$ RL & 2 & 163 & 667 & 200 & 73.8 (1030) & 3,909 & 39 \\
\quad Alignment SFT $\rightarrow$ RL & 1 & 167 & 683 & 215 & 76.3 (1065) & 3,185 & 21 \\
\quad Alignment SFT $\rightarrow$ RL & 2 & 167 & 691 & 203 & 76.0 (1061) & 3,039 & 14 \\
\quad \ours{} with $\beta{=}0$ (control) & 1 & 162 & 689 & 206 & 75.7 (1057) & 3,742 & 23 \\
\quad \ours{} with $\beta{=}0$ (control) & 2 & 160 & 691 & 206 & 75.7 (1057) & 3,676 & 15 \\
\quad \ours{}: alignment SFT $\rightarrow$ attention RL & 1 & 166 & 688 & 216 & 76.6 (1070) & 3,472 & 32 \\
\quad \ours{}: alignment SFT $\rightarrow$ attention RL & 2 & 163 & 697 & 209 & 76.6 (1069) & 3,610 & 31 \\
\midrule
\multicolumn{8}{@{}l}{\textit{Inference-time methods on the base model}} \\*
\quad ViCrop (rel-att, layer 23) & 1 & 161 & 685 & 211 & 75.7 (1057) & 4,122 & 20 \\
\quad ViCrop (rel-att, layer 23) & 2 & 163 & 682 & 216 & 76.0 (1061) & 4,337 & 18 \\
\quad VCD & 1 & 166 & 682 & 203 & 75.3 (1051) & 4,612 & 101 \\
\quad ZoomEye & 1 & 174 & 634 & 199 & 72.1 (1007) & 5,371 & 92 \\
\quad Fixed-grid zoom (2$\times$2 tiles) & 1 & 164 & 668 & 203 & 74.1 (1035) & 4,725 & 40 \\
\quad Zoom-Refine (two passes) & 1 & 165 & 658 & 201 & 73.4 (1024) & 8,232 & 13 \\
\quad Self-Refine & 1 & 160 & 645 & 203 & 72.2 (1008) & 7,994 & 6 \\
\quad Length-aware visual logit bias, $\beta{=}0.25$ & 1 & 160 & 666 & 208 & 74.1 (1034) & 4,751 & 48 \\
\quad Length-aware visual logit bias, $\beta{=}0.5$ & 1 & 155 & 670 & 206 & 73.9 (1031) & 5,013 & 42 \\
\quad Length-aware visual logit bias, $\beta{=}1$ & 1 & 161 & 668 & 207 & 74.2 (1036) & 5,100 & 54 \\
\midrule
\multicolumn{8}{@{}l}{\textit{Head-guided zoom (inference-time crops from attention maps)}} \\*
\quad Base + localization-head crop & 1 & 171 & 675 & 195 & 74.6 (1041) & 4,276 & 22 \\
\quad Alignment SFT $\rightarrow$ RL + localization-head crop & 1 & 170 & 695 & 195 & 75.9 (1060) & 2,857 & 7 \\
\quad Alignment SFT $\rightarrow$ RL + layer-23 head-average crop & 1 & 163 & 695 & 204 & 76.1 (1062) & 2,894 & 9 \\
\quad Alignment SFT $\rightarrow$ RL + both crops & 1 & 168 & 694 & 212 & 76.9 (1074) & 2,959 & 9 \\
\quad Alignment SFT $\rightarrow$ RL + both crops & 2 & 169 & 691 & 204 & 76.2 (1064) & 2,984 & 8 \\
\midrule
\multicolumn{8}{@{}l}{\textit{Released checkpoints (own backbones; raw prompt, single pass, no tools)}} \\*
\quad ZwZ-4B & 1 & 171 & 649 & 196 & 72.8 (1016) & 185 & 0 \\
\quad ZwZ-7B & 1 & 159 & 594 & 181 & 66.9 (934) & 31 & 0 \\
\quad TreeVGR-7B & 1 & 159 & 586 & 187 & 66.8 (932) & 60 & 0 \\
\quad DeepEyes-7B & 1 & 136 & 587 & 159 & 63.2 (882) & 45 & 0 \\
\quad VGPO-7B & 1 & 143 & 576 & 163 & 63.2 (882) & 41 & 0 \\
\quad VPPO-7B & 1 & 144 & 574 & 162 & 63.0 (880) & 56 & 0 \\
\quad Reflection-V-7B & 1 & 141 & 568 & 165 & 62.6 (874) & 78 & 0 \\
\quad NoisyRollout-7B & 1 & 149 & 565 & 154 & 62.2 (868) & 45 & 0 \\
\quad Qwen2.5-VL-7B & 1 & 149 & 545 & 159 & 61.1 (853) & 44 & 0 \\
\quad LLaVA-OneVision-72B & 1 & 147 & 534 & 167 & 60.7 (848) & 9 & 0 \\
\quad Remember-R1-7B & 1 & 140 & 554 & 145 & 60.1 (839) & 47 & 0 \\
\quad Pixel-Reasoner-7B & 1 & 143 & 544 & 130 & 58.5 (817) & 119 & 1 \\
\quad LLaVA-OneVision-7B & 1 & 139 & 524 & 152 & 58.4 (815) & 2 & 0 \\
\quad InternVL3-8B & 1 & 122 & 489 & 166 & 55.7 (777) & 68 & 0 \\
\midrule
\multicolumn{8}{@{}l}{\textit{Released checkpoints with an explicit-think prompt (prompt-sensitivity check)}} \\*
\quad TreeVGR-7B & 1 & 157 & 594 & 190 & 67.4 (941) & 65 & 0 \\
\quad VPPO-7B & 1 & 152 & 570 & 171 & 64.0 (893) & 79 & 0 \\
\quad VGPO-7B & 1 & 143 & 574 & 155 & 62.5 (872) & 44 & 0 \\
\quad DeepEyes-7B & 1 & 148 & 559 & 160 & 62.1 (867) & 58 & 0 \\
\quad Reflection-V-7B & 1 & 137 & 565 & 143 & 60.5 (845) & 170 & 0 \\
\quad Remember-R1-7B & 1 & 129 & 567 & 135 & 59.5 (831) & 65 & 0 \\
\quad Pixel-Reasoner-7B & 1 & 112 & 526 & 114 & 53.9 (752) & 130 & 0 \\
\end{longtable}

\begin{table}[!ht]
\centering
\small
\setlength{\tabcolsep}{4.4pt}
\begin{tabular}{@{}lrrrrrrl@{}}
\toprule
Method & V$^*$ & HR-4K & Tree & Overall & Tokens & Trunc.\ (\%) & $\Delta$ vs.\ base [95\% CI] \\
\midrule
Qwen3.5-4B (base) & 78.0 & 75.8 & 42.7 & 66.5 & 7,144 & 19.1 & --- \\
ViCrop (rel-att) & 79.1 & 79.1 & 41.2 & 68.1 & 6,498 & 16.8 & $+1.6$ [$-$0.6, 4.0] \\
GRPO & 73.3 & 76.5 & 42.0 & 66.0 & 7,737 & 20.9 & $-0.4$ [$-$2.3, 1.5] \\
DAPO & 75.4 & 79.0 & 44.2 & 68.4 & 5,830 & 14.8 & $+1.9$ [$-$0.1, 4.0] \\
EASE & 73.8 & 74.8 & 44.0 & 65.7 & 8,002 & 21.8 & $-0.8$ [$-$2.7, 1.2] \\
\midrule
CE SFT $\rightarrow$ RL & 80.6 & 82.1 & 45.7 & 71.3 & 4,864 & 10.4 & $+4.9$ [2.7, 7.1] \\
Alignment SFT $\rightarrow$ RL & 82.7 & 83.5 & 49.4 & 73.5 & 3,691 & 6.4 & $+7.0$ [4.8, 9.3] \\
\bottomrule
\end{tabular}
\caption{\textbf{Greedy decoding} (temperature 0, same budget and scoring; one decode per model). $\Delta$: paired difference to the base model with 95\% CIs. Alignment SFT$\rightarrow$RL leads CE SFT$\rightarrow$RL by $+2.1$ points (95\% CI [0.4, 4.0]).}
\label{tab:t0}
\end{table}

\section{Full-Width Figures}
\label{app:figures}

\FloatBarrier

\end{document}